\documentclass[lettersize,journal]{IEEEtran}

\usepackage{cite}
\usepackage{amsmath,amssymb,amsfonts}
\usepackage{graphicx}
\usepackage{subcaption}
\usepackage{pgfplots}
\usepackage{comment}
\usepackage{multirow}
\usepackage{array}
\usepackage{xcolor}      

\usepackage{algorithm}
\usepackage{algpseudocode}
\usepackage[hidelinks]{hyperref} 
\usepackage[capitalise]{cleveref} 

\makeatletter
\def\@IEEEBIOskipN{1.5\baselineskip}
\makeatother

\newtheorem{proposition}{Proposition}

\title{Communication-Aware UAV Spatio-Temporal Relay Cooperation and Planning in Complex Urban Environments}

\author{Haoyu~Zhang, Yingjie~He, Zehao~Zhuo, Xiqun~Chen, and~Zhenyu~Gao%
\thanks{This research was supported by the Research Grants Council (RGC) of Hong Kong through the Collaborative Research Fund (CRF) under project number C6078-25G and by The National Natural Science Foundation of China (NSFC) under Grants 72525009 and 72431009 (Corresponding authors: Xiqun Chen and Zhenyu Gao).}%
\thanks{Haoyu Zhang, Yingjie He, and Zhenyu Gao are with the Department of Mechanical and Aerospace Engineering, The Hong Kong University
of Science and Technology, Hong Kong, China (e-mail: zhenyugao@ust.hk).}%
\thanks{Zehao Zhuo and Xiqun Chen are with the Institute of Intelligent Transportation Systems, College of Civil Engineering and Architecture, Zhejiang University, Hangzhou 310058, China (e-mail: chenxiqun@zju.edu.cn).}%
}

\begin{document}

\maketitle

\begin{abstract}

Unmanned aerial vehicles (UAVs) are increasingly deployed in low-altitude urban airspace for delivery, emergency response, and other applications. However, UAV mobility across spatially heterogeneous urban communication environments can lead to time-varying communication outages, disrupting mission execution and compromising flight safety. To address this challenge, this paper proposes a Spatio-Temporal Relay Cooperation and Planning (ST-RCP) framework that leverages delivery UAVs as opportunistic communication relays for disconnected UAVs without requiring additional communication infrastructure. The framework predicts communication blind-spot events over a finite planning horizon, constructs a helper–event compatibility graph, and jointly optimizes relay selection, obstacle-aware trajectory replanning, and communication-aware scheduling. Multiple planners are evaluated to investigate the trade-offs among communication coverage, computational cost, and energy consumption. Experiments based on real-world traffic scenarios compare the performance of ST-RCP against baseline methods. In the peak scenario, ST-RCP improves descent-phase communication coverage by 29.93\% compared with the base-station baseline and by 27.38\%, 22.02\%, and 15.20\% compared with configurations using one, two, and three dedicated relay UAVs, respectively. Process flexibility analysis demonstrates that predictive task-aware relaying expands the pool of feasible relay resources and improves communication continuity while maintaining the feasibility of delivery missions. These findings demonstrate the potential of trajectory-enabled UAV coordination to enhance the safety and resilience of urban aerial operations.

\end{abstract}

\section{Introduction}

Unmanned aerial vehicles (UAVs) are playing an increasingly important role in low-altitude urban air mobility, with growing applications in logistics delivery, emergency response, and search and rescue \cite{gao2024noise,10803914}. Reliable communication is essential for these operations, as command and control signals, traffic management information, and mission-related data must be exchanged between UAVs and ground infrastructure to ensure safe and efficient operations \cite{greenfeld2019uam,8470897}. Communication disruptions can pose significant risks to flight safety, as the loss of command and control connectivity may compromise UAV operations and increase the risk of accidents.

\begin{figure}[t]
    \centering
    \includegraphics[width=0.95\linewidth]{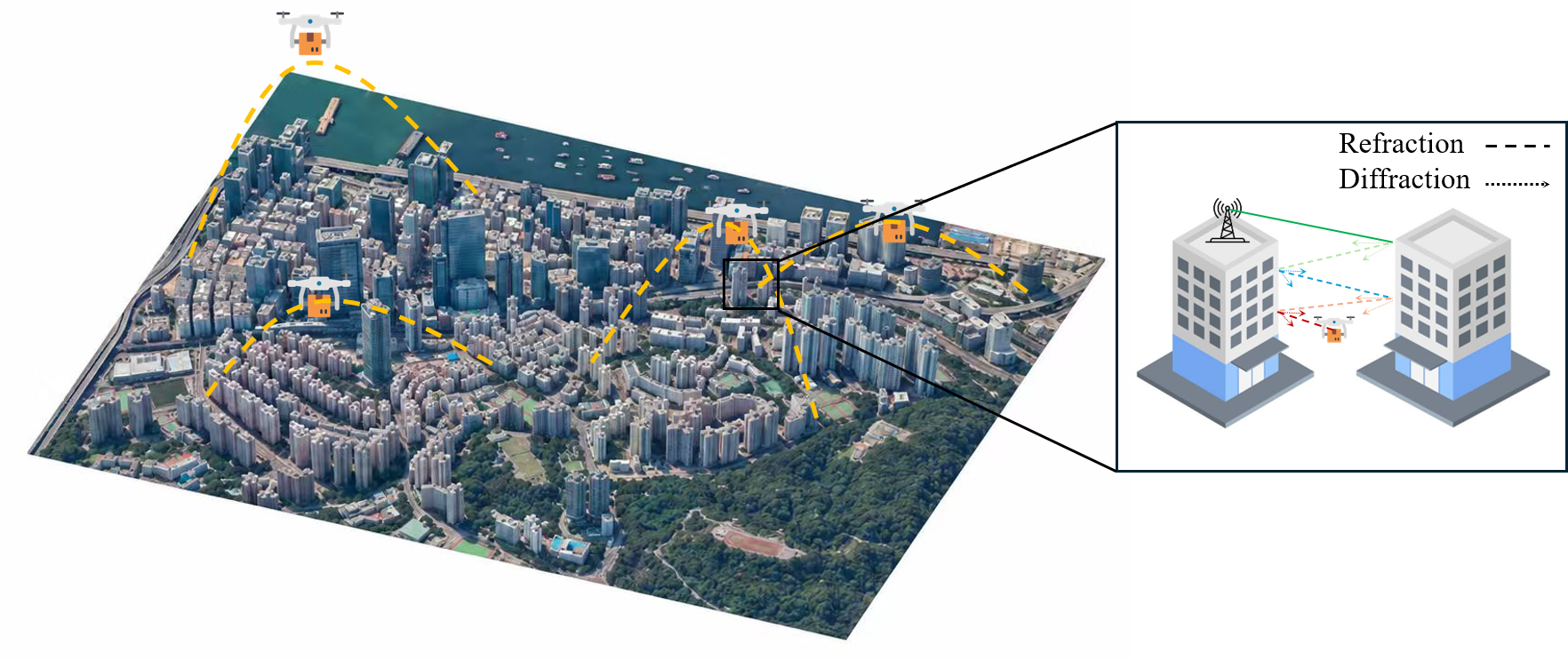}
    \caption{Urban canyons disrupt communication links during urban delivery task.}
    \label{fig:spatial_blindspot}
\end{figure}

Maintaining reliable connectivity, however, is particularly challenging in dense urban environments. High-rise buildings can obstruct line-of-sight (LoS) communication links and cause signal reflection, diffraction, and scattering, leading to substantial spatial variations in received signal strength and localized communication coverage gaps, particularly in urban canyons and areas obstructed by buildings \cite{8709739,11162132}, as illustrated in Fig.~\ref{fig:spatial_blindspot}.

Simply increasing base station density may not fully resolve these challenges. UAVs may establish strong propagation links with multiple neighboring base stations simultaneously, potentially exacerbating co-channel interference and limiting the coverage and throughput gains of network densification \cite{8692749}. Furthermore, additional base stations may not eliminate coverage gaps caused by signal blockage from urban structures \cite{11162132}. UAV connectivity is also inherently trajectory dependent, particularly when mission locations are uncertain. As a UAV moves through a spatially heterogeneous radio environment, its received signal quality varies over time, causing it to repeatedly enter and leave communication coverage regions \cite{d1d2325ba27041e9ab61322119b00720}. Therefore, reliable urban UAV communication depends not only on the spatial distribution of radio coverage but also on the spatiotemporal interaction between UAV trajectories and communication conditions.

Several approaches have been proposed to mitigate intermittent UAV connectivity. One representative approach is to establish flying ad hoc networks (FANETs), in which neighboring UAVs exchange data through multi-hop communication links. However, the high mobility of UAVs leads to rapidly changing network topologies, making it difficult for conventional routing protocols to maintain reliable end-to-end communication paths \cite{ullah2025spectral}. Moreover, network-wide flooding of routing messages can generate substantial redundant transmissions, increasing channel contention and routing overhead \cite{kilicarslan2025broadcast}. An alternative approach is to deploy UAV-mounted base stations or dedicated relay UAVs to restore or extend communication coverage. Although such aerial relays can effectively improve connectivity, they require additional aerial resources and incur substantial deployment and operational costs \cite{cong2025multi,yang2025multiobjective}. More recently, studies have explored the use of delivery UAVs as communication relays by adjusting their flight routes and introducing communication waypoints to support multi-hop connectivity \cite{soya2025flight}. Such studies demonstrate the potential of leveraging the mobility of task-performing UAVs to support communication without relying exclusively on dedicated relay platforms. However, existing approaches largely plan communication topologies and delivery routes in advance, rather than dynamically adapting to time-varying communication outages arising from task uncertainty. This limitation motivates a more dynamic planning framework that treats future communication blind spots as time-dependent service events and predictively reserves and reroutes available task-performing UAVs to provide relay assistance while satisfying communication, flight safety, obstacle avoidance, energy, and delivery constraints.

To address this research gap, this paper proposes a spatio-temporal relay cooperation and planning (ST-RCP) framework to enable self-organizing UAV communication in complex urban environments. Unlike conventional approaches that rely on dedicated relay platforms, ST-RCP leverages the mobility of UAVs already performing delivery missions, allowing available UAVs to temporarily serve as opportunistic communication relays while maintaining the feasibility of their original tasks. Furthermore, rather than optimizing communication locations along predetermined routes, ST-RCP models communication outages as time-varying service demands and coordinates relay resources based on both spatial reachability and temporal availability. Specifically, future communication blind-spot events and their associated attributes are predicted over a finite planning horizon. For each event, ST-RCP identifies candidate delivery UAVs capable of providing relay assistance while satisfying their original mission constraints. A helper–event compatibility graph is then constructed to capture spatial, temporal, and communication feasibility. This formulation transforms relay selection from a static waypoint placement problem into a constrained spatio-temporal resource allocation problem. Based on the selected helpers, smooth, obstacle-aware relay trajectories are subsequently generated using B\'ezier curves.

The proposed framework is evaluated in terms of communication performance, operational efficiency, and computational cost. Beyond conventional coverage metrics, this study introduces a process flexibility perspective to characterize the structural adaptability of the task-performing UAV relay system. This analysis complements conventional performance evaluations by examining not only the effectiveness of relay strategies under given scenarios but also the system's ability to accommodate varying communication demands, resource contention, and fleet sizes. The main contributions of this work are summarized as follows:
\begin{enumerate}
    \item A spatio-temporal relay cooperation and planning (ST-RCP) framework is proposed to enable task-performing UAVs to opportunistically provide communication relay assistance without requiring additional infrastructure.
    \item Communication blind-spot recovery is formulated as a predictive helper–event assignment and constrained trajectory planning problem, enabling time-dependent, event-driven relay resource allocation.
    \item A process flexibility analysis framework is introduced to quantify the scalability and adaptability of UAV relay systems, characterizing their ability to maintain feasible relay options under increasing traffic demand and resource contention.
\end{enumerate}

\section{Literature Review}\label{sec:lit}
\subsection{UAV Communication and Relay Strategies}

FANETs establish connectivity through UAV-to-UAV multi-hop links, with recent studies using clustering, reinforcement learning, link-stability prediction, and selective forwarding to improve routing under high mobility. Ullah \emph{et al.}\cite{ullah2025spectral} combine spectral clustering with reinforcement learning, while Li and Sun\cite{li2025prediction} adapt route discovery using predicted link stability. Kılıçarslan \emph{et al.}\cite{kilicarslan2025broadcast} show that unrestricted rebroadcast can cause redundant transmissions and channel contention, and He \emph{et al.}\cite{he2025fanet} jointly optimize UAV trajectories, topology, and transmit power to improve throughput. Despite these advances, FANETs remain sensitive to rapid topology changes and link instability, while frequent neighbor discovery and routing updates can introduce substantial communication overhead and limit scalability in dense or highly dynamic UAV networks.

Dedicated UAV relays and UAV-mounted base stations offer another approach to improving aerial connectivity. Cong \emph{et al.}\cite{cong2025multi} jointly optimize scheduling, transmit power, and trajectories for multi-UAV relaying, while Yang \emph{et al.}\cite{yang2025multiobjective} optimize relay population, placement, communication performance, and propulsion energy. Although such dedicated aerial resources can enhance coverage and throughput, they introduce additional deployment, scheduling, and energy costs, which increase as the network scales. Tsai and Huang\cite{tsai2025implementation} further demonstrate the practical feasibility of UAV-based relaying using a 3GPP-compatible onboard communication module. More recently, Soya \emph{et al.}\cite{soya2025flight} employ delivery UAVs as multi-hop communication nodes by introducing communication waypoints into predefined routes, showing that task-performing UAVs can also support communication. However, these methods remain largely route-oriented and do not explicitly model time-varying communication outages, temporal relay demand, or competition among limited relay resources. These limitations motivate predictive and event-driven relay coordination that jointly considers relay assignment, temporal availability, and trajectory feasibility.
\subsection{Communication- and Obstacle-Aware UAV Trajectory Planning in Urban Environments}

In urban low-altitude environments, UAV trajectory planning must jointly consider communication and mission constraints. Zhong \emph{et al.}\cite{zhong2024joint} combine radio and environmental maps with D3QN to reduce communication interruptions, mission time, and obstacle risk, while related work uses radio maps and deep reinforcement learning to maintain connectivity in three-dimensional flight space \cite{liu2024deep3d}. These studies incorporate communication constraints into trajectory optimization but primarily focus on maintaining the connectivity of individual UAVs, with limited consideration of coordinating UAVs to provide time-bounded relay assistance to others.

Urban buildings also require explicit geometric planning. Hu \emph{et al.} \cite{zhang2025layered} model buildings as cuboid obstacles in a layered visibility graph and add maneuvering protection spots to reduce collision risk. Banik \emph{et al.}\cite{banik2025performance} compare PRM, RRT*, and EST in Gazebo using a three-dimensional occupancy grid. Li \emph{et al.} \cite{li2026energy} jointly optimize NOMA power and a three-dimensional relay trajectory under several constraints. These results reinforce a dual-constraint view of the city: buildings restrict physically feasible flight space and simultaneously obstruct radio propagation. Obstacle avoidance and communication planning should therefore not be treated as independent post-processing steps.

Multi-UAV trajectory planning further requires safe operation. Existing studies have incorporated conflict avoidance, delivery efficiency, and energy constraints into UAV routing. Pang \emph{et al.} \cite{PANG2022103666} formulate multi-UAV 4D conflict resolution as a mixed-integer nonlinear optimization problem that jointly coordinates scheduling, speed adjustment, and rerouting to generate conflict-free trajectories. Chen \emph{et al.} \cite{chen2026hotspot} construct conflict-free and energy-aware routes using reachable spatio-temporal regions under uncertain low-altitude traffic demand, while related cargo-delivery studies incorporate communication reliability into pickup-and-delivery routing \cite{chen2024cargo}. These studies demonstrate that safety, connectivity, and operational efficiency can be incorporated into UAV trajectory design, but they generally optimize predefined missions or trajectories without simultaneously considering dynamically arriving communication-service demands.

\subsection{Summary}

Existing studies have investigated FANET routing, dedicated UAV relays, and communication-aware trajectory planning. However, these approaches face limitations associated with dynamic network topologies, additional relay resource requirements, and an emphasis on predefined missions or individual UAV connectivity. The joint consideration of time-varying communication outages, competition for limited relay resources, and feasible relay trajectory replanning remains insufficiently explored. To address these limitations, ST-RCP integrates predictive, event-driven relay coordination with constraint-aware trajectory replanning to enhance communication continuity in dynamic urban environments.

\section{Methodology}
\label{sec:methodology}

\begin{figure*}[t]
    \centering
    \includegraphics[width=0.9\textwidth]
    {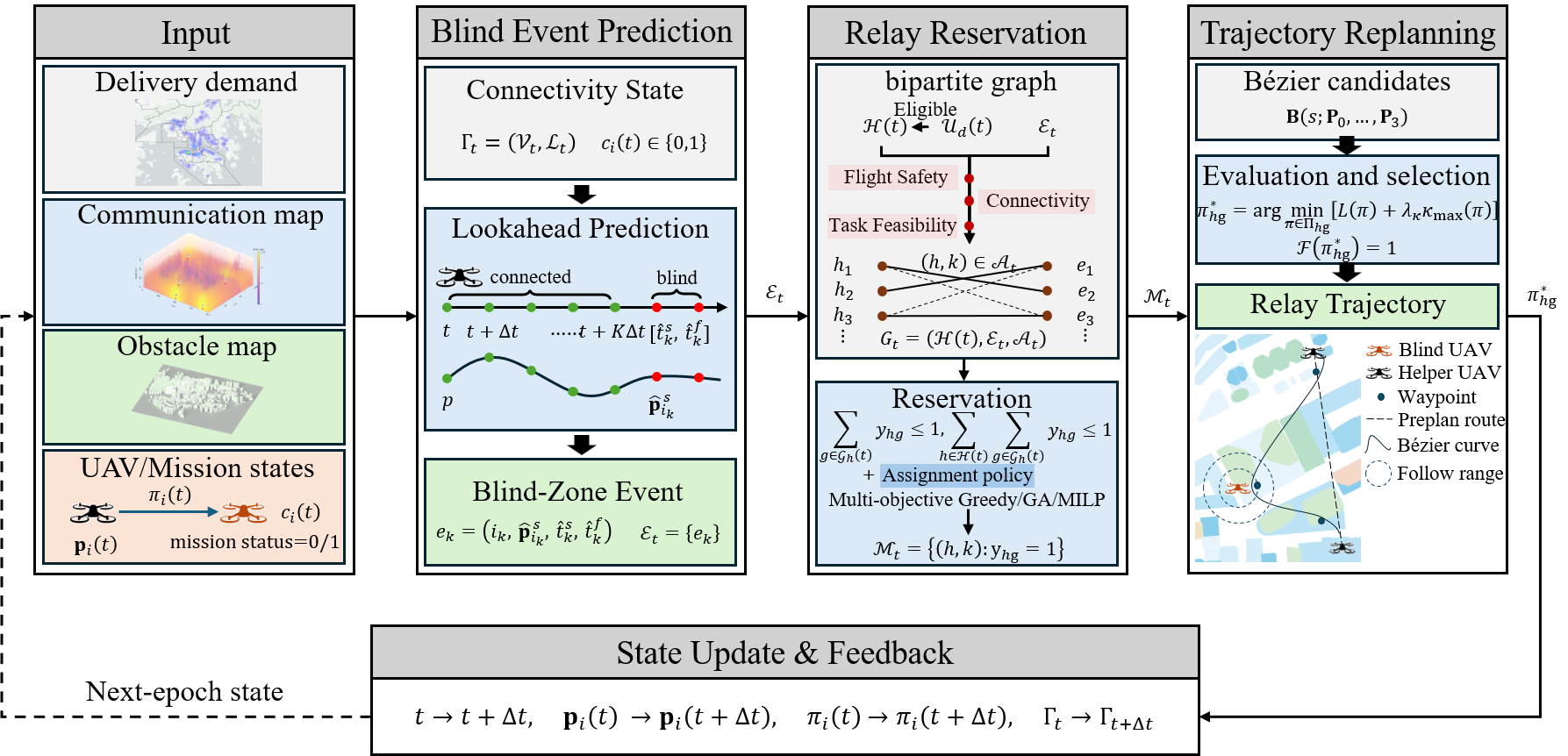}
    \caption{Overall closed-loop framework of ST-RCP. Delivery demand, communication and obstacle maps, and UAV/mission states provide the system inputs. Future connectivity degradation is converted into time-bounded blind-zone events, which are matched with eligible helper UAVs through a helper--event bipartite graph and relay cooperation. Selected helper--group assignments are then processed by constraint-aware B\'ezier trajectory replanning to generate feasible relay trajectories. The updated UAV and mission states are fed back to the next planning epoch, enabling continuous spatio-temporal relay coordination.}
    \label{fig:strrp_framework}
\end{figure*}

The ST-RCP framework is designed to address communication support requirements across a wide range of urban low-altitude operations. Among emerging applications, UAV-based food delivery services have already been deployed at scale in several regions, providing a representative operational setting in which communication reliability, UAV mobility, obstacle avoidance, and service continuity must be jointly considered. We therefore adopt UAV-based food delivery as the application scenario to illustrate the proposed framework and to conduct the subsequent experimental evaluation and analysis. Fig.~\ref{fig:strrp_framework} presents the overall framework of ST-RCP. The framework takes delivery demand, the communication and obstacle maps, and the current UAV and mission states as inputs. Based on the current connectivity state, future communication losses are predicted and converted into time-bounded blind zone events $\mathcal E_t$. Eligible delivery UAVs and predicted events are then organized into a helper--event bipartite graph $G_t$, from which relay assignments $\mathcal M_t$ are selected and reserved. For each selected helper--group assignment, constraint-aware B\'ezier trajectory replanning generates a feasible relay trajectory $\pi_{hg}^{*}$. The resulting UAV and mission states are subsequently fed back to the next planning epoch, forming a closed-loop relay coordination process.

\subsection{Problem Formulation}
\label{subsec:system_problem}

The ST-RCP system consists of fixed ground base stations, a fleet of task-performing delivery UAVs, and a three-dimensional urban environment with buildings. Let $\mathcal{U}_{d}(t)$ denote the set of active delivery UAVs at time $t$, $\mathcal{S}$ the set of base stations, and $\mathcal{B}$ the set of building volumes. Each UAV follows a time-dependent delivery trajectory $\pi_i(t)$ with position $\mathbf{p}_i(t)\in\mathbb{R}^{3}$.

Buildings jointly affect mobility and communication by restricting collision-free flight space and obstructing line-of-sight propagation. The instantaneous communication state is represented by a time-varying graph $\Gamma_t=(\mathcal{V}_t,\mathcal{L}_t)$, where $\mathcal{V}_t=\mathcal{U}_{d}(t)\cup\mathcal{S}$. An edge in $\mathcal{L}_t$ represents a feasible communication link satisfying the adopted propagation, communication-range, and building-aware line-of-sight conditions, assuming orthogonal channel allocation between BS--UAV backhaul and UAV--UAV relay access links to avoid co-channel self-interference.

Let $\mathcal{E}_{t}$ denote the set of predicted blind-zone events at time $t$. Each event $k$ is represented as
\begin{equation}
e_k =
\left(
i_k,\,
\widehat{\mathbf{p}}_{i_k}^{\,\mathrm{s}},\,
\widehat{t}_k^{\mathrm{s}},\,
\widehat{t}_k^{\mathrm{f}}
\right),
\label{eq:outage_event}
\end{equation}
where $i_k$ is the affected UAV, $\widehat{\mathbf{p}}_{i_k}^{\,\mathrm{s}}$ is its predicted position at outage onset, and $\widehat{t}_k^{\mathrm{s}}$ and $\widehat{t}_k^{\mathrm{f}}$ denote the predicted start and end times of the outage. Thus, a trajectory-dependent communication loss is formulated as a time-bounded relay-service request.

For each event, an available delivery UAV may temporarily deviate from its nominal trajectory to serve as a helper, provided that it can reach the required relay region within the event window and satisfy the communication, obstacle, safety, energy, and mission constraints.

\subsection{Predictive Blind-Zone Event Modeling}
\label{subsec:predictive_events}

ST-RCP uses a finite lookahead horizon to identify imminent losses of communication reachability before they occur. For UAV $i$, future positions are sampled along its current trajectory as $\widehat{\mathbf{p}}_i(t+\ell\Delta t),\ell=0,\ldots,K_H,$ where $\Delta t$ is the simulation time step and $K_H$ is the number of lookahead samples.

Let $\widehat{c}_i(t+\ell\Delta t)$ denote the predicted reachability state at a sampled future position. The lookahead trigger is
\begin{equation}
\operatorname{Trigger}_i(t)
=
\mathbf{1}
\left[
\exists \ell\in\{0,\ldots,K_H\}:
\widehat{c}_i(t+\ell\Delta t)=0
\right].
\label{eq:lookahead_trigger}
\end{equation}
When the trigger is activated, the predicted loss of reachability is encoded as an event $e_k$ using Eq.~\eqref{eq:outage_event}. The event therefore specifies not only the affected UAV, but also when and where relay support is expected to become necessary. This representation provides the temporal and spatial information required for subsequent helper--event compatibility evaluation and relay reservation.

\subsection{Helper--Event Compatibility and Relay Cooperation}
\label{subsec:compatibility}

At time $t$, the available helper set is denoted by $\mathcal{H}(t)\subseteq\mathcal{U}_{d}(t)$. A delivery UAV $h$ is included in the eligible helper set $\mathcal{H}(t)$ if it satisfies the following conditions: \textit{(1) Active mission:} The UAV has departed and has not completed its delivery mission. \textit{(2) Helper availability:} The UAV is not currently providing relay assistance to another UAV. \textit{(3) Rescue exclusion:} The UAV is not itself receiving relay assistance. \textit{(4) Cooldown eligibility:} The prescribed cooldown following its previous detour has elapsed, allowing it to undertake another relay task.

For each helper--event pair $(h,k)$, candidate relay positions are iteratively evaluated around the affected UAV's predicted location using predefined spatial offsets. Each position is screened against SNR, three-dimensional line-of-sight, communication range, and altitude constraints. For feasible positions, candidate transit trajectories and arrival times are generated, forming a set of relay options
\begin{equation}
\mathcal{O}_{hk}
=
\left\{
o=(\mathbf{r}_{hk},\pi_{hk},\tau_{hk})
\right\},
\end{equation}
where $\mathbf{r}_{hk}$ is the candidate relay position, $\pi_{hk}$ is the corresponding helper trajectory, and $\tau_{hk}$ is the estimated transit time. For spatially proximate events with overlapping service intervals, a common relay position is further considered if it can simultaneously satisfy the communication and safety constraints of all active targets. Such events can be grouped and served by a single helper, whereas events without a feasible shared position are handled independently as single-event groups.

The compatibility between helper $h$ and event $k$ is defined as
\begin{equation}
a_{hk}(t)=
\begin{cases}
1, & h\in\mathcal{H}(t),\; \exists o\in\mathcal{O}_{hk}:\phi(o)=1,\\
0, & \text{otherwise},
\end{cases}
\label{eq:compatibility}
\end{equation}
where
\begin{equation}
\phi(o)=\chi_{\mathrm{r}}(o)\,\chi_{\mathrm{c}}(o)\,\chi_{\mathrm{b}}(o)\,\chi_{\mathrm{s}}(o)\,\chi_{\mathrm{d}}(o),
\label{eq:feasibility_predicate}
\end{equation}
and each indicator equals one when its corresponding condition is satisfied and zero otherwise.

Here, $\chi_{\mathrm{r}}$, $\chi_{\mathrm{c}}$, $\chi_{\mathrm{b}}$, $\chi_{\mathrm{s}}$, and $\chi_{\mathrm{d}}$ denote local-radius, communication, obstacle, separation-safety, and delivery-task screening, respectively. These indicators provide a pre-dispatch feasibility filter; the selected relay trajectory is subsequently subjected to the time-indexed execution checks in Eq.~\eqref{eq:trajectory_feasibility}. The task screen uses the estimated delivery-delay ratio
$q_h=(2\tau_{hk}+\tau_h^{\mathrm{follow}})/T_h^{\mathrm{route}}$,
and requires $q_h\leq q_{\max}$. This bound limits disturbance to the helper's original delivery mission, while delivery deadlines are evaluated separately in the mission-level metrics.

The feasible helper--event relationships form a time-varying bipartite graph
\begin{equation}
G_t=
\left(
\mathcal{H}(t),
\mathcal{E}_t,
\mathcal{A}_t
\right),
\qquad
\mathcal{A}_t=
\{(h,k):a_{hk}(t)=1\}.
\label{eq:compatibility_graph}
\end{equation}

Based on the compatibility graph, events with overlapping service intervals and feasible common relay positions are grouped for shared assistance. Let $\mathcal{G}_h(t)$ denote the set of feasible relay-service groups for helper $h$, including single-event ($|g|=1$) and shared-event ($|g|>1$) groups, where each group $g\in\mathcal{G}_h(t)$ is associated with a validated relay position $\mathbf{r}_g$. The binary decision variable $y_{hg}\in\{0,1\}$ indicates whether helper $h$ is assigned to group $g$.
The assignment constraints are
\begin{equation}
\sum_{g\in\mathcal{G}_h(t)}y_{hg}\leq 1,
\qquad \forall h\in\mathcal{H}(t),
\label{eq:helper_capacity}
\end{equation}
and
\begin{equation}
\sum_{h\in\mathcal{H}(t)}
\sum_{\substack{g\in\mathcal{G}_h(t)\\k\in g}}
y_{hg}\leq 1,
\qquad \forall k\in\mathcal{E}_t.
\label{eq:event_assignment}
\end{equation}

These constraints allow a single helper to serve multiple events within a feasible shared group, while preventing duplicate event assignments and conflicting helper reservations. Relay assignments are enforced through helper availability and reservation states during execution. Algorithm~\ref{alg:helper_selection} summarizes the compatibility evaluation, group construction, and relay-reservation procedure.

\begin{algorithm}[t]
\caption{Helper--Event Compatibility Graph Construction and Relay Cooperation}
\label{alg:helper_selection}
\begin{algorithmic}[1]
\Require Active delivery UAV set $\mathcal U_d(t)$, predicted event set $\mathcal E_t$, obstacle map, radio map, task constraints, delay bound $q_{\max}$, departure shift bound $\bar{\delta}_h$
\Ensure Reserved helper--group matching $\mathcal M_t$

\State Optimize departure shifts $\boldsymbol{\delta}$ for pre-departure UAVs via Eq.~\eqref{eq:flexibility_scheduling}
\State Initialize eligible helper set $\mathcal H(t)\gets\emptyset$, edge set $\mathcal A_t\gets\emptyset$
\ForAll{$u \in \mathcal U_d(t)$}
    \If{$u$ is available for relay service at time $t$}
        \State $\mathcal H(t)\gets \mathcal H(t)\cup\{u\}$
    \EndIf
\EndFor

\ForAll{$k\in\mathcal E_t$}
    \ForAll{$h\in\mathcal H(t)$}
        \If{$h\neq i_k$}
            \State Evaluate feasibility predicate $a_{hk}(t)$
            \If{$a_{hk}(t)=1$}
                \State $\mathcal A_t\gets\mathcal A_t\cup\{(h,k)\}$
            \EndIf
        \EndIf
    \EndFor
\EndFor

\State Construct compatibility graph $G_t=(\mathcal H(t),\mathcal E_t,\mathcal A_t)$
\If{$\mathcal A_t=\emptyset$}
    \State \Return $\mathcal M_t=\emptyset$
\EndIf

\State Construct feasible service groups $\{\mathcal G_h(t)\}_{h\in\mathcal H(t)}$ from $G_t$
\State Apply assignment policy subject to \eqref{eq:helper_capacity}--\eqref{eq:event_assignment} to determine $\{y_{hg}\}$
\State Obtain reserved helper--group matching $\mathcal M_t=\{(h,g):y_{hg}=1\}$
\ForAll{$(h,g)\in\mathcal M_t$}
    \State Mark helper $h$ as reserved for service group $g$
\EndFor

\State \Return $\mathcal M_t$
\end{algorithmic}
\end{algorithm}

The reserved helper--group matching $\mathcal M_t$ is determined by applying an assignment policy to the feasible service groups derived from $G_t$. To support predictive relay coordination, ST-RCP maintains a cache of blind-zone events and adopts an event-driven reservation mechanism. Relay assignments are updated upon changes in mission states, reservation commitments, dispatch failures, or relay-service completion, with periodic checks ensuring timely updates. For each assigned group, the scheduler estimates the latest feasible helper departure time, enabling advance reservation and positioning before communication outages occur. This mechanism reduces redundant replanning while preserving timely relay assistance. The helper-selection policies evaluated in this study are described in Section~\ref{subsec:eval_methods}.

\subsection{Constraint-Aware Relay Trajectory Replanning}
\label{subsec:trajectory_replanning}

Following relay reservation, a helper's original delivery trajectory is retained for subsequent restoration. To reduce redundant computation, initial reservations use estimated travel times. When a reservation $(h,g)\in\mathcal{M}_t$ enters the locking window before the latest feasible departure time, candidate relay trajectories are generated from the helper's predicted departure state to the group relay position $\mathbf{r}_g$. The actual trajectory travel time is then used to refine the departure schedule, while infeasible reservations trigger helper reselection. The transit trajectory is parameterized using a cubic B\'ezier curve:
\begin{equation}
\mathbf{B}(s)
=
\sum_{j=0}^{3}
\binom{3}{j}(1-s)^{3-j}s^{j}\mathbf{P}_j,
\qquad s\in[0,1],
\label{eq:bezier_curve}
\end{equation}
where $\mathbf{P}_0$ is the helper's predicted departure position and $\mathbf{P}_3=\mathbf{r}_g$ is the selected relay position for group $g$. The intermediate control points are generated from the helper's current direction and a bounded set of local offsets.

Let $\Pi_{hg}$ denote the set of admissible candidate trajectories generated for helper $h$ and service group $g$. Among these candidates, the implemented curve generator selects a trajectory according to
\begin{equation}
\pi_{hg}^{*}
=
\arg\min_{\pi\in\Pi_{hg}}
\left[
L(\pi)
+
\lambda_{\kappa}\kappa_{\max}(\pi)
\right],
\label{eq:trajectory_objective}
\end{equation}
where $L(\pi)$ is the curve length and $\kappa_{\max}(\pi)$ is the maximum sampled curvature. Equation~\eqref{eq:trajectory_objective} therefore favors shorter and smoother relay trajectories.

Candidate trajectories are subjected to detailed time-indexed feasibility checks before final selection and dispatch, complementing the pair-level screening in Eq.~\eqref{eq:feasibility_predicate}. Let $\mathcal{T}_{hg}^{\mathrm{tr}}$ denote the helper's transit interval to $\mathbf{r}_g$, and let $\mathcal{T}_{hk}^{\mathrm{relay}}$ denote the active relay-service interval for member event $k\in g$. The trajectory feasibility indicator is
\begin{equation}
\mathcal{F}(\pi)
=
\mathbf{1}
\left[
\begin{aligned}
&\min_{t \in \mathcal{T}_{hg}^{\mathrm{tr}}}
d_{\mathrm{obs}}\!\left(\mathbf{p}_h(t)\right)
\ge d_{\mathrm{clr}},
\\
&\min_{\substack{
t \in \mathcal{T}_{hg}^{\mathrm{tr}} \\
j \neq h
}}
\left\|
\mathbf{p}_h(t)-\mathbf{p}_j(t)
\right\|
\ge d_{\mathrm{safe}},
\\
&\min_{t \in \mathcal{T}_{hk}^{\mathrm{relay}}}
\gamma_{hk}(t)
\ge \gamma_{\mathrm{th}},
\qquad
\forall k \in g,
\\
&\mathrm{LoS}_{hk}(t)=1,
\qquad
\forall t \in \mathcal{T}_{hk}^{\mathrm{relay}},\;\forall k \in g
\end{aligned}
\right],
\label{eq:trajectory_feasibility}
\end{equation}
where $d_{\mathrm{obs}}(\mathbf{p}_h(t))$ denotes the distance from the helper to the nearest obstacle, $d_{\mathrm{clr}}$ is the required obstacle clearance, and $d_{\mathrm{safe}}$ is the minimum inter-UAV separation. $\gamma_{hk}(t)$ denotes the relay-link SNR between helper $h$ and the affected UAV $i_k$ associated with member event $k\in g$, and $\gamma_{\mathrm{th}}$ is the required SNR threshold. The last condition requires line-of-sight availability throughout each member event's relay interval. Only trajectories satisfying all applicable execution constraints are retained as feasible candidates. For shared relay service ($|g|>1$), the communication and safety conditions are verified for every member event $k\in g$ over its respective active service interval $\mathcal{T}_{hk}^{\mathrm{relay}}$, rather than requiring all target links to be active simultaneously before their scheduled start times.

To justify the discrete candidate-generation scheme in Eq.~\eqref{eq:trajectory_objective}, we establish the asymptotic convergence of the sampled trajectory optimum and its geometric perturbation bound under control-point refinement.

\begin{proposition}[Convergence under Nested Control-Point Refinement]
\label{prop:bezier_convergence}
Fix the helper--group assignment $(h,g)$, predicted departure state $\mathbf{P}_0$, and relay position $\mathbf{r}_g$, and use the same time-indexed feasibility model at every refinement level. Let $\mathcal{D}$ be the continuous domain of intermediate control-point pairs $\mathbf{x}=(\mathbf{P}_1,\mathbf{P}_2)$ permitted by the local-offset rule. Write $\pi_{\mathbf{x}}$ for the trajectory generated by Eq.~\eqref{eq:bezier_curve} from $\mathbf{x}$ and let $\mathcal{S}=\{\mathbf{x}\in \mathcal{D}:\mathcal{F}(\pi_{\mathbf{x}})=1\}$, with $\mathcal{F}$ evaluated as in Eq.~\eqref{eq:trajectory_feasibility} for the applicable group constraints. Define $J(\mathbf{x})=L(\pi_{\mathbf{x}})+\lambda_\kappa\kappa_{\max}(\pi_{\mathbf{x}})$ as in Eq.~\eqref{eq:trajectory_objective} and let $J^\star=\inf_{\mathbf{x}\in \mathcal{S}}J(\mathbf{x})$, assumed finite. Let $\mathcal{X}_n\subseteq \mathcal{D}$ be finite candidate sets with $\mathcal{X}_n\subseteq \mathcal{X}_{n+1}$ and $\bigcup_n \mathcal{X}_n$ dense in $\mathcal{D}$. Suppose that for every $\varepsilon>0$, there is a point $\bar{\mathbf{x}}\in \mathcal{S}$ with $J(\bar{\mathbf{x}})<J^\star+\varepsilon$ at which $J$ is continuous and which has a neighborhood relative to $\mathcal{D}$ contained in $\mathcal{S}$. Then the discrete optimum $J_n=\min_{\mathbf{x}\in \mathcal{X}_n\cap \mathcal{S}}J(\mathbf{x})$, with $J_n=+\infty$ when $\mathcal{X}_n\cap \mathcal{S}=\varnothing$, is nonincreasing and converges to $J^\star$.
\end{proposition}

\begin{IEEEproof}
Since $\mathcal{X}_n\cap \mathcal{S}\subseteq \mathcal{X}_{n+1}\cap \mathcal{S}\subseteq \mathcal{S}$, we have $J^\star\le J_{n+1}\le J_n$. For any $\varepsilon>0$, choose $\bar{\mathbf{x}}$ as above. By continuity and local feasibility, a relative neighborhood of $\bar{\mathbf{x}}$ consists entirely of feasible pairs with objective values below $J^\star+\varepsilon$. Density guarantees that some $\mathcal{X}_N$ intersects this neighborhood. By nesting, the resulting candidate is retained for every $n\ge N$, so $J^\star\le J_n<J^\star+\varepsilon$. Hence $J_n\to J^\star$.
\end{IEEEproof}

The cubic B\'ezier representation in Eq.~\eqref{eq:bezier_curve} provides a direct geometric condition to verify local feasibility. Since the endpoints $\mathbf{P}_0$ and $\mathbf{P}_3=\mathbf{r}_g$ are fixed, if each intermediate control point in $\mathbf{x}=(\mathbf{P}_1,\mathbf{P}_2)$ is perturbed from $\bar{\mathbf{x}}$ by at most $r$ in the Euclidean norm, the maximum spatial deviation between the resulting trajectories satisfies
\begin{equation}
\sup_{s\in[0,1]}\|\mathbf{B}_{\mathbf{x}}(s)-\mathbf{B}_{\bar{\mathbf{x}}}(s)\|_2
\le \sup_{s\in[0,1]} 3s(1-s)r
=\frac{3r}{4}.
\label{eq:bezier_perturbation}
\end{equation}

Consequently, a uniform obstacle-clearance margin $m>0$ beyond $d_{\mathrm{clr}}$ is guaranteed to be preserved for any control-point perturbation satisfying $r\le 4m/3$. The same bound applies to the inter-UAV separation margin $d_{\mathrm{safe}}$ when the time parameterization and surrounding UAV trajectories are fixed, as well as to the link-quality and line-of-sight margins over the service intervals. For fixed curvature-sampling parameters, the continuity of $J(\mathbf{x})$ follows directly from the smoothness of polynomial curves, provided that the reference curve has a nonvanishing first derivative at each sample point.

After reaching the relay position, for a single-event assignment ($|g|=1$), the helper regulates its escort speed according to
\begin{equation}
v_h^{\mathrm{cmd}}(t)
=
\operatorname{clip}_{[0,v_h^{\max}]}
\!\left(
v_i(t)+k_p[d_{hi}(t)-d^{\star}(t)]
\right),
\label{eq:escort_speed}
\end{equation}
where $i=i_k$ denotes the affected UAV, $d_{hi}(t)=\|\mathbf p_h(t)-\mathbf p_i(t)\|_2$ is their separation, and $d^{\star}(t)=d_0+\tau_g v_i(t)$ is the desired following distance. The clipping operator bounds the commanded speed between zero and $v_h^{\max}$. Here, $d_0$ is the base following distance, $\tau_g$ is the desired time gap, and $k_p$ is the distance-error gain. For shared relay service ($|g|>1$), the helper maintains the validated common relay position $\mathbf{r}_g$ while monitoring communication and safety constraints for all active member events. The helper resumes its saved delivery trajectory when all member events in $g$ complete their blind-zone traversal, relay assistance becomes infeasible, or the maximum service duration is reached. Algorithm~\ref{alg:trajectory_replanning} summarizes the relay-trajectory generation, feasibility validation, and execution process for a reserved helper--group assignment $(h,g)$.

\begin{algorithm}[t]
\caption{Constraint-Aware Relay Trajectory
Replanning and Execution}
\label{alg:trajectory_replanning}
\begin{algorithmic}[1]

\Require Reserved helper--group assignment $(h,g)$,
predicted departure state, relay position
$\mathbf r_g$, obstacle and radio maps

\Ensure Feasible relay trajectory $\pi_{hg}^{*}$
or failure flag

\State Generate candidate B\'ezier trajectories
from the predicted departure position
to $\mathbf r_g$

\State Initialize feasible set
$\Pi_{hg}\gets\emptyset$

\ForAll{candidate trajectory $\pi$}
    \State Evaluate trajectory feasibility
    using Eq.~\eqref{eq:trajectory_feasibility}

    \If{$\mathcal F(\pi)=1$}
        \State $\Pi_{hg}\gets\Pi_{hg}\cup\{\pi\}$
    \EndIf
\EndFor

\If{$\Pi_{hg}=\emptyset$}
    \State Trigger helper reselection
    \State \Return failure
\EndIf

\State Select the feasible trajectory
$\pi_{hg}^{*}$ minimizing length and curvature via Eq.~\eqref{eq:trajectory_objective}

\State Refine the latest feasible departure
time using the validated trajectory duration

\State Dispatch helper $h$ along $\pi_{hg}^{*}$

\State Maintain safe standby until
the scheduled relay service begins

\If{$|g|=1$}
    \While{the event is active and relay
    assistance remains feasible}
        \State Update escort speed using
        Eq.~\eqref{eq:escort_speed}
    \EndWhile
\Else
    \While{shared relay service remains active}
        \State Maintain the validated
        common relay position $\mathbf r_g$
        \State Check communication and safety
        constraints for each active event $k\in g$
        \State Update active service members
    \EndWhile
\EndIf

\State Restore the original delivery trajectory
\State Release the helper reservation
\State \Return $\pi_{hg}^{*}$

\end{algorithmic}
\end{algorithm}

\subsection{Process-Flexibility-Aware Departure Scheduling}
\label{subsec:flexibility}

We propose bounded departure time adjustments to configure the helper--event flexibility graph before relay assignment. For an eligible, not-yet-departed helper $h$, its departure time is shifted to $t_h^{0}+\delta_h$, where $0\leq\delta_h\leq\bar{\delta}_h$. The adjustment changes its predicted availability and position, thereby modifying the compatibility edges while preserving order-release, delivery-deadline, and safety constraints.

The departure adjustments are selected through
\begin{equation}
\max_{\boldsymbol{\delta}}
\left\{
\min_{\omega\in\Omega}
\nu\!\left(G_{\omega}(\boldsymbol{\delta})\right)
-\lambda\sum_h\delta_h
\right\},
\label{eq:flexibility_scheduling}
\end{equation}
where $\Omega$ includes nominal operation and specified helper-unavailability scenarios. The quantity $\nu(G)$ is the maximum number of events that can be jointly served under compatibility, cooperation, and trajectory-conflict constraints. It therefore extends ordinary bipartite matching to temporally feasible assignments.
The coefficient $\lambda$ penalizes departure delay.

This formulation configures alternative helper--event relationships to improve service capacity under resource unavailability. Departure schedules are evaluated over a finite candidate set, with compatibility and trajectory feasibility recomputed for each schedule.

\section{Numerical Experiment}
\label{sec:evaluation}

We conduct a simulation study to evaluate the effectiveness, applicability, and underlying mechanisms of ST-RCP in realistic urban delivery networks. The evaluation compares ST-RCP with representative communication strategies in terms of communication reliability, operational cost, and computational efficiency, and further examines its sensitivity to system scale, infrastructure density, relay conditions, and planning constraints. In addition, process flexibility analysis is used to investigate how helper--event compatibility, temporal slack, and relay-resource availability affect the robustness and scalability of the proposed framework. We evaluate the proposed framework from three perspectives: (i) overall performance compared with representative baseline methods, particularly in terms of communication coverage and operational cost; (ii) process flexibility, including helper compatibility, resource redundancy, and arrival-time margin; and (iii) sensitivity to key system and operating conditions.

\subsection{Experimental Environment and Data}
\label{subsec:eval_environment}

\subsubsection{Urban Environment and Communication Setup}

\begin{figure*}[t]
    \centering

    \begin{subfigure}[t]{0.27\linewidth}
        \centering
        \includegraphics[width=\linewidth]
        {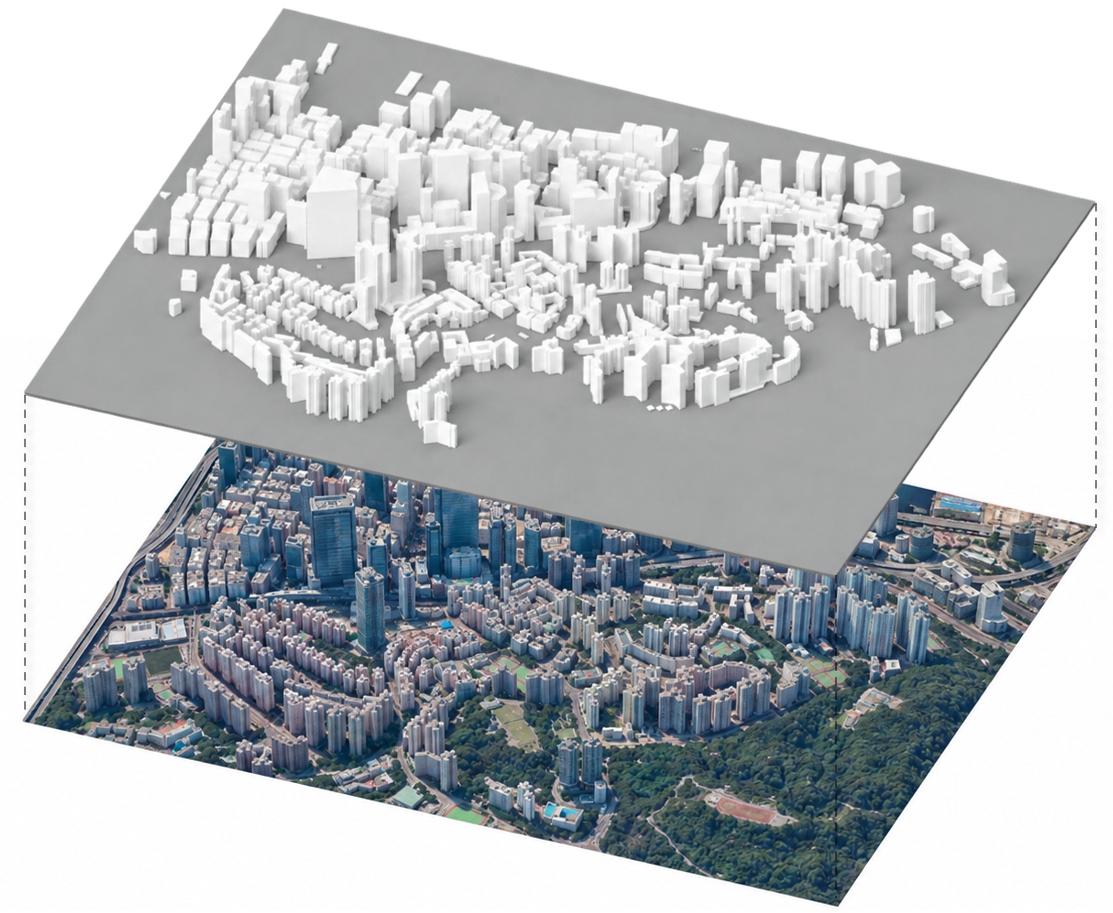}
        \caption{Reconstructed urban environment.}
        \label{fig:urban_reconstruction}
    \end{subfigure}
    \hfill
    \begin{subfigure}[t]{0.33\linewidth}
        \centering
        \includegraphics[width=\linewidth]
        {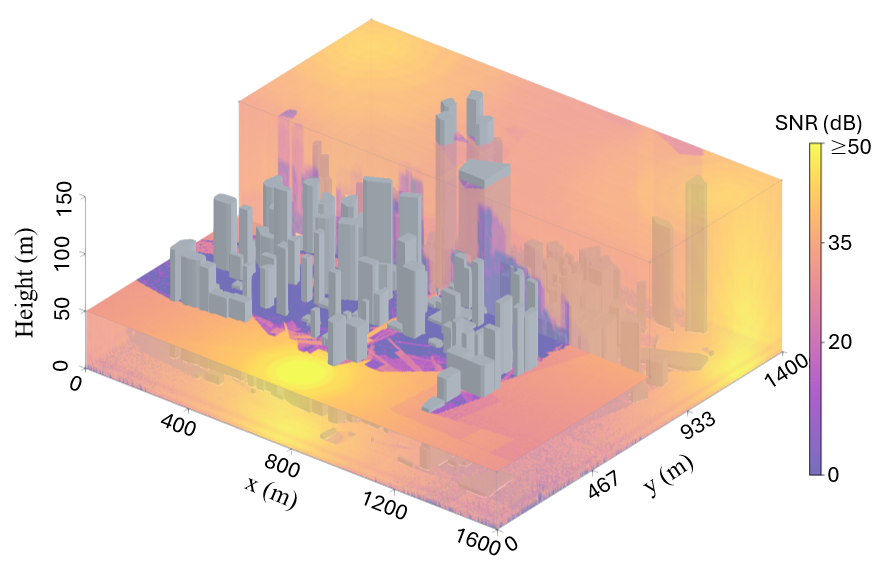}
        \caption{The three-BS deployment.}
        \label{fig:snr_3bs}
    \end{subfigure}
    \hfill
    \begin{subfigure}[t]{0.33\linewidth}
        \centering
        \includegraphics[width=\linewidth]
        {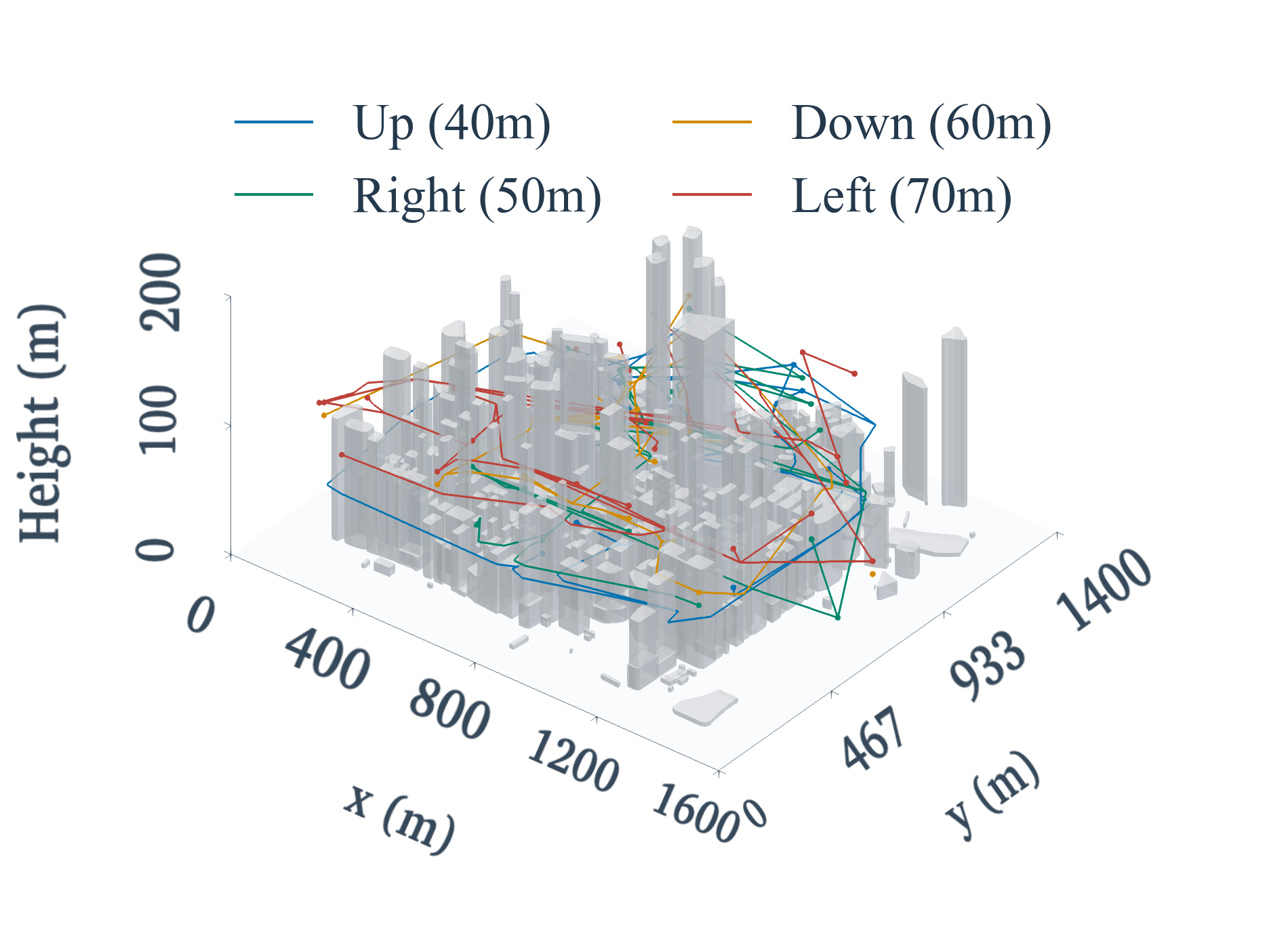}
        \caption{Preplanned UAV delivery trajectories.}
        \label{fig:trajectory_environment}
    \end{subfigure}

    \caption{Urban simulation environment and communication maps:
    (a) reconstructed three-dimensional environment of Kwun Tong, Hong Kong;
    (b) Stepped cut of the 3-base-station SNR field; and
    (c) Preplanned UAV delivery trajectories over four altitude layers.}
    \label{fig:urban_and_snr}
\end{figure*}

The numerical experiments are conducted in an urban environment reconstructed from OpenStreetMap data for Kwun Tong, Hong Kong. Kwun Tong is a densely built industrial district characterized by dense building blocks and pronounced urban canyon effects. Fig.~\ref{fig:urban_reconstruction} illustrates the reconstructed three-dimensional urban environment used in the experiments.

\paragraph{Radio-map generation}
NVIDIA Sionna RT is employed to perform three-dimensional ray tracing in the reconstructed urban environment. For a base station (BS) $m\in\mathcal{S}$ located at $\mathbf{b}_m$ and a spatial location $\mathbf{p}=(x,y,z)$, let $L_{\mathrm{ch}}(\mathbf{b}_m,\mathbf{p})$ denote the channel propagation gain obtained through ray tracing. Based on the link budget model, the received signal power is calculated as
\begin{equation}
P_m(\mathbf{p})
=
P_t + G_t
+ L_{\mathrm{ch}}(\mathbf{b}_m,\mathbf{p})
+ G_r ,
\end{equation}
where $P_t$ denotes the BS transmit power, and $G_t$ and $G_r$ denote the BS and UAV antenna gains, respectively. In the experiments, we set $P_t=40\,\mathrm{dBm}$, $G_t=20\,\mathrm{dB}$, carrier frequency $f_c=39\,\mathrm{GHz}$, channel bandwidth $B=4\times10^{8}\,\mathrm{Hz}$, receiver noise figure $N_F=5\,\mathrm{dB}$, and $G_r=8\,\mathrm{dB}$.

\paragraph{Base station placement}
We use a greedy method to select a fixed number of ground BS sites \cite{he2026coplanning}. At each step, we choose the candidate site that yields the largest marginal increase in map-wide communication coverage. The resulting three-dimensional Signal-to-noise ratio (SNR) field with three BSs is illustrated in Fig.~\ref{fig:snr_3bs}.

\paragraph{Connectivity and relay feasibility}
The terrestrial network covers location $\mathbf{p}$ if
$Q_{\mathcal{S}}(\mathbf{p}) \geq \gamma_{\mathrm{th}} = 25\,\mathrm{dB}$,
where $Q_{\mathcal{S}}(\mathbf{p})$ is the highest BS--UAV SNR recorded in the radio map at $\mathbf{p}$ across the deployed ground base stations $\mathcal{S}$. The basic connectivity graph limits UAV--UAV links to $200\,\mathrm{m}$. During helper dispatch and trajectory execution, relay links must also satisfy the applicable SNR and line-of-sight requirements.

\subsubsection{Delivery data and trajectory scenarios}

The order loader is based on real-world Deliveroo order data collected on 8 October 2023, including pickup, drop-off, restaurant, and order-time information. The original geographic coordinates are transformed into the local simulation coordinate system using the affine mapping implemented in the scenario loader. For the main experiments, delivery UAVs follow preplanned trajectories consisting of a cruise segment followed by a final descent segment. The resulting delivery trajectories are embedded in the reconstructed three-dimensional urban environment, as illustrated in Fig.~\ref{fig:trajectory_environment}. To reduce potential flight conflicts, the cruise trajectories are assigned to multiple altitude layers at $40$, $50$, $60$, and $70\,\mathrm{m}$.

\begin{figure}[t]
    \centering

    \begin{subfigure}[t]{0.48\linewidth}
        \centering
        \includegraphics[width=\linewidth]{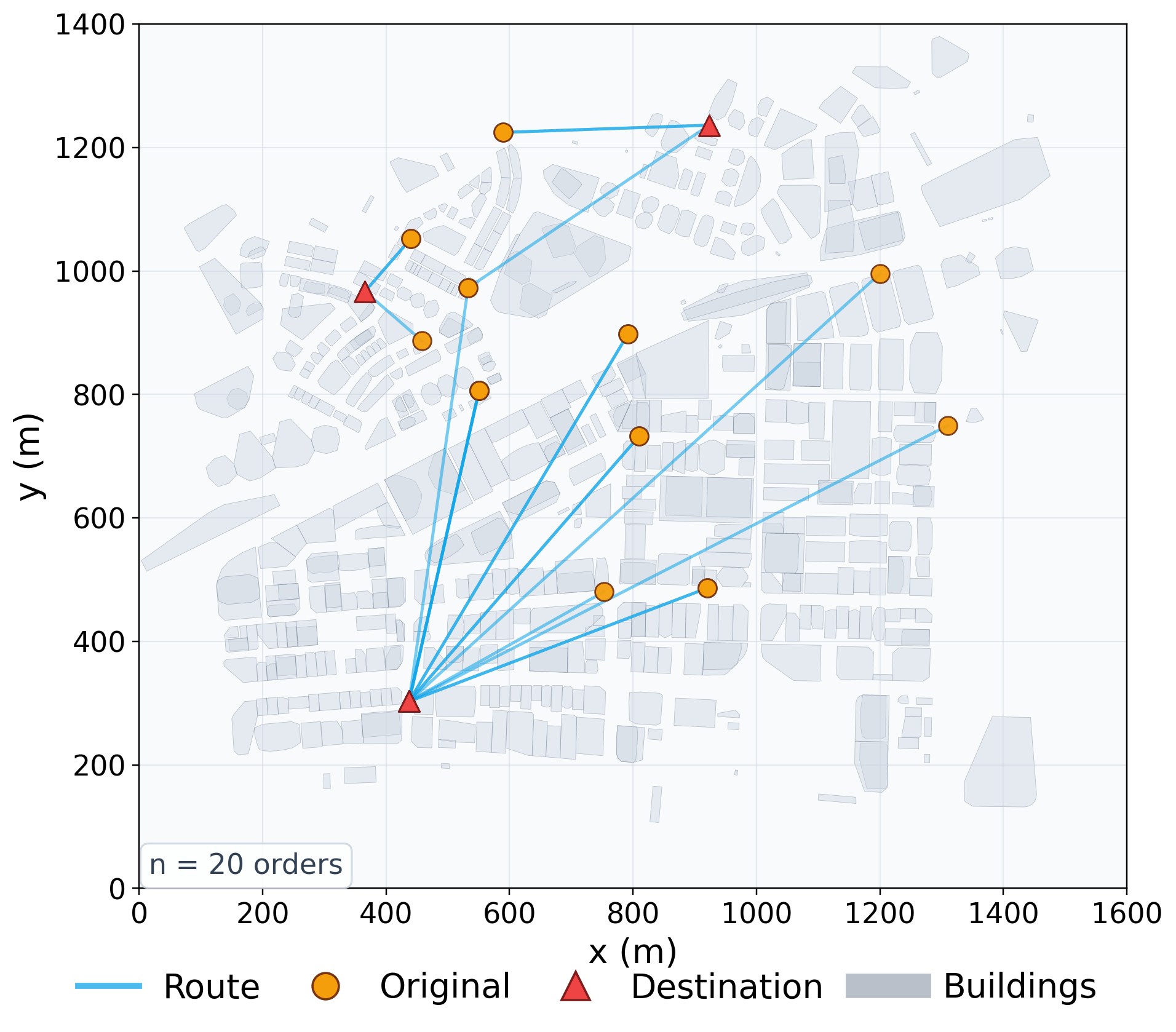}
        \caption{Off-peak scenario (20 orders).}
        \label{fig:off_peak}
    \end{subfigure}
    \hfill
    \begin{subfigure}[t]{0.48\linewidth}
        \centering
        \includegraphics[width=\linewidth]{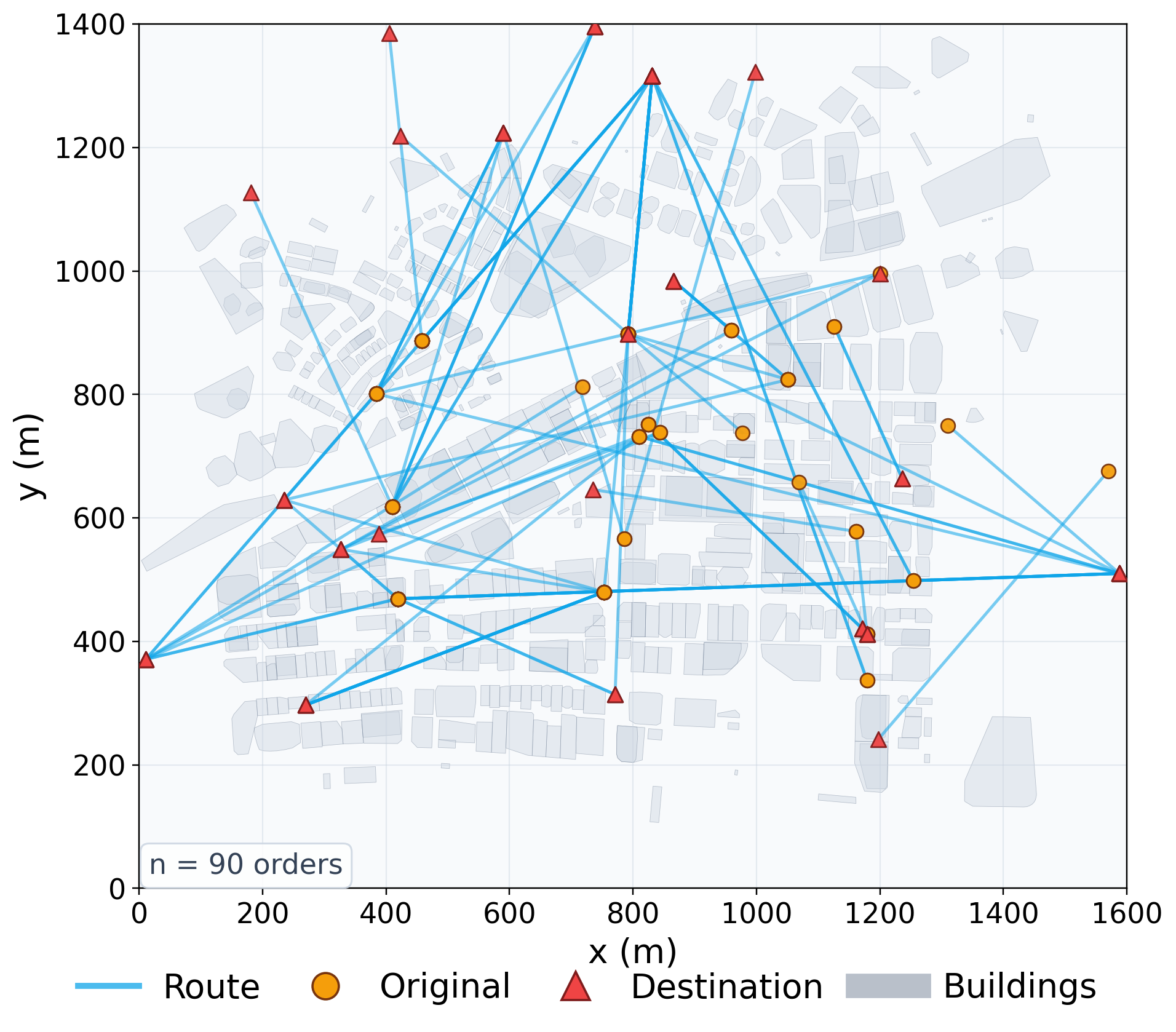}
        \caption{Peak scenario (90 orders).}
        \label{fig:peak}
    \end{subfigure}

    \caption{Delivery scenarios used in the main evaluation.}
    \label{fig:order_distribution}
\end{figure}

The main evaluation considers two one-hour delivery scenarios derived from real-world order records, representing off-peak and peak periods with 20 and 90 orders, respectively. Fig.~\ref{fig:order_distribution} illustrates the spatial distributions of the two delivery scenarios, representing low- and high-demand operating conditions while preserving the spatial and temporal characteristics of the original order data. Additional controlled scenarios are constructed for scalability, sensitivity, and process flexibility analyses to independently examine the effects of fleet size, helper availability, and the spatiotemporal distribution of communication blind-spot events. Other key experimental parameters are summarized in Table~\ref{tab:common_settings}.

\begin{table}[t]
\centering
\caption{Experimental parameter settings.}
\label{tab:common_settings}
\small
\begin{tabular}{ll}
\hline
{\bf Parameter} & {\bf Setting} \\
\hline
Horizontal map size & $1600\times1400\,\mathrm{m}$ \\
Voxel/radio resolution & $1\,\mathrm{m}$ \\
Radio altitude range & $0$--$150\,\mathrm{m}$ \\
Main base-station count & 5 \\
Sensitivity base-station count & 3 \\
Base-station altitude & $10\,\mathrm{m}$ \\
SNR threshold & $25\,\mathrm{dB}$ \\
Cruise/descent speed & $3/1\,\mathrm{m/s}$ \\
Simulation step & $0.5\,\mathrm{s}$ \\
Predictive lookahead & $600\,\mathrm{s}$ \\
Predictive replanning interval & $60\,\mathrm{s}$ \\
Default helper radius & $300\,\mathrm{m}$ \\
Default maximum delay ratio & $0.5$ \\
Dedicated relay count & 1--3 in sensitivity experiments \\
Maximum departure shift $\bar{\delta}_h$ & $30\,\mathrm{s}$ \\
Departure-shift penalty $\lambda$ & $0.05\,\mathrm{s}^{-1}$ \\
\hline
\end{tabular}
\end{table}

\subsection{Compared Methods and ST-RCP Selectors}
\label{subsec:eval_methods}

The comparison includes five groups of methods.

\begin{enumerate}
    \item \textbf{Base Station:}
    Delivery UAVs communicate directly with terrestrial base stations without UAV-to-UAV relaying, representing the conventional infrastructure-only baseline.

    \item \textbf{Dense Base Station:}
    Additional terrestrial base stations are deployed using the same coverage-maximization procedure to improve communication coverage without UAV-to-UAV relay assistance.
    
    \item \textbf{FANET:}
    Delivery UAVs form a multi-hop ad hoc network to maintain connectivity with terrestrial infrastructure based on the current network topology.

    \item \textbf{Dedicated Relay UAV:}
    Dedicated communication UAVs provide relay support to poorly connected delivery UAVs. Reactive and predictive variants are considered, with the latter repositioning relays using predicted blind-zone information.

    \item \textbf{ST-RCP helper-selection strategies:}
    Greedy, genetic algorithms (GA) and a global mixed-integer linear programming (MILP) benchmark are employed for helper selection. To evaluate the impact of optimization objectives, four communication-oriented criteria are considered: target coverage (TC), network coverage (NC), coverage per energy (CE), and coverage per distance (CD), formulated as
    \begin{equation}
    \label{eq:selection_objectives}
    \begin{aligned}
        \mathrm{TC}:&
        \max_{\mathcal{M}_t} \sum_{(h,g)\in\mathcal{M}_t} c_{h,g},
        \quad
        \mathrm{NC}:
        \max_{\mathcal{M}_t} \sum_{(h,g)\in\mathcal{M}_t} g_{h,g},\\
        \mathrm{CE}:&
        \max_{\mathcal{M}_t}
        \frac{\sum_{(h,g)\in\mathcal{M}_t} g_{h,g}}
        {\epsilon+\sum_{(h,g)\in\mathcal{M}_t} E_{h,g}},\\
        \mathrm{CD}:&
        \max_{\mathcal{M}_t}
        \frac{\sum_{(h,g)\in\mathcal{M}_t} g_{h,g}}
        {\epsilon+\sum_{(h,g)\in\mathcal{M}_t} L_{h,g}},
    \end{aligned}
    \end{equation}
    where $\mathcal{M}_t$ denotes the selected feasible helper--group assignments; $c_{h,g}$ and $g_{h,g}$ represent the recovered outage duration and network-wide coverage gain for group $g$, respectively; $E_{h,g}$ and $L_{h,g}$ denote the additional energy consumption and detour distance; and $\epsilon>0$ prevents division by zero. All objectives share identical candidate-generation procedures and feasibility constraints.
\end{enumerate}

\subsection{Evaluation Metrics}
\label{subsec:eval_metrics}

The evaluation considers four categories of metrics: communication performance, operational cost, computational performance, and process flexibility.

\textbf{Communication performance.}
Let $c_i(t)\in\{0,1\}$ denote whether UAV $i$ is connected at time $t$, $\mathcal{T}_i$ its active-flight epochs, and $\mathcal{T}_i^{\mathrm{des}}\subseteq\mathcal{T}_i$ its descent-phase epochs. The total and descent-phase coverage ratios are

\begin{equation}
C_{\mathrm{total}}
=
\frac{\sum_{i\in\mathcal U}\sum_{t\in\mathcal T_i}c_i(t)}
{\sum_{i\in\mathcal U}|\mathcal T_i|},
\qquad
C_{\mathrm{des}}
=
\frac{\sum_{i\in\mathcal U}\sum_{t\in\mathcal T_i^{\mathrm{des}}}c_i(t)}
{\sum_{i\in\mathcal U}|\mathcal T_i^{\mathrm{des}}|}.
\label{eq:coverage_metrics}
\end{equation}

The descent-phase coverage is reported separately because communication loss near the delivery destination is particularly relevant to the relay mechanism considered in this work.

\textbf{Operational cost.}
To evaluate the propulsion-energy cost of UAV operations, we adopt a speed-dependent power model motivated by the rotary-wing UAV propulsion-energy modeling in \cite{zeng2019energy}. The total system energy consumption is calculated over discrete simulation steps as
\begin{equation}
E_{\mathrm{total}}
=
\sum_{i\in\mathcal{U}}
\sum_{t\in\mathcal{T}_i}
\left(p_0+c\,v_i^2(t)\right)\Delta t,
\label{eq:energy_metrics}
\end{equation}
where $v_i(t)$ is UAV's instantaneous speed, $p_0=170\,\mathrm{W}$ represents the baseline propulsion power, $c=2\,\mathrm{W\,s^2/m^2}$ is the speed-dependent power coefficient, and $\Delta t=0.5\,\mathrm{s}$ is the simulation time step.

\textbf{Process flexibility.}
Inspired by process flexibility in industrial engineering~\cite{jordan1995principles}, this study evaluates the ability of UAV relay systems to accommodate varying communication demands through alternative feasible assignments represented by the helper--event compatibility graph. At evaluation epoch $t$, let $G_t=(\mathcal H(t),\mathcal E_t,\mathcal A_t)$ denote the helper--event compatibility graph, where $\mathcal A_t$ contains all feasible helper--event pairs. We characterize process flexibility using three metrics: compatibility-edge count, event redundancy, and temporal margin. Let
$\mathcal T_E=\{t\in\mathcal T:|\mathcal E_t|>0\}$ and
$\mathcal T_A=\{t\in\mathcal T:|\mathcal A_t|>0\}$.
The metrics are defined as
\begin{equation}
\begin{aligned}
\bar{A}
&=
\frac{1}{|\mathcal T|}
\sum_{t\in\mathcal T}|\mathcal A_t|,
\\
\bar{R}
&=
\frac{1}{|\mathcal T_E|}
\sum_{t\in\mathcal T_E}
\frac{|\mathcal A_t|}{|\mathcal E_t|},
\\
\bar{M}
&=
\frac{1}{|\mathcal T_A|}
\sum_{t\in\mathcal T_A}
\frac{1}{|\mathcal A_t|}
\sum_{(h,k)\in\mathcal A_t}
\left(t_k^{\mathrm f}-t_{hk}^{\mathrm{arr}}\right).
\end{aligned}
\label{eq:flexibility_metrics}
\end{equation}

Here, $\bar{A}$ measures the average number of feasible helper--event relationships, $\bar{R}$ measures the average number of alternative helpers available per active event, and $\bar{M}$ measures the average remaining service-time margin after a feasible helper reaches the relay position. Larger values indicate greater relay-resource availability, redundancy, and temporal flexibility, respectively. These metrics characterize the system's potential capacity to accommodate relay requests, independent of the number of relays actually executed.

\begin{table*}[t]
    \centering
    \caption{Performance comparison in the 20-order off-peak scenario.}
    \label{tab:offpeak_comparison}
    \small
    \begin{tabular}{lcccc}
        \hline
        \textbf{Method} &
        \textbf{Total Coverage} &
        \textbf{Descent Coverage} &
        \textbf{Total Energy (J)} &
        \textbf{Additional Energy (J)} \\
        \hline
        Base Station        & 82.50\% & 24.62\% & 55187.44 & 0.00 \\
        FANET               & 89.15\% & 53.64\% & 55187.44 & 0.00 \\
        Dense Base Station (8)  & 91.63\% & 78.01\% & 55187.44 & 0.00 \\
        Dedicated Relay UAV (3) & \textbf{95.63\%} & \textbf{81.86\%} & 62135.56 & 6948.12 \\
        ST-RCP              & 92.27\% & 66.86\% & \textbf{56304.26} & \textbf{1116.82} \\
        \hline
    \end{tabular}
\end{table*}

\begin{table*}[t]
    \centering
    \caption{Performance comparison in the 90-order peak scenario.}
    \label{tab:peak_comparison}
    \small
    \begin{tabular}{lcccc}
        \hline
        \textbf{Method} &
        \textbf{Total Coverage} &
        \textbf{Descent Coverage} &
        \textbf{Total Energy (J)} &
        \textbf{Additional Energy (J)} \\
        \hline

        Base Station
        & 76.62\%
        & 51.73\%
        & 346691.87
        & 0.00 \\

        FANET
        & 93.72\%
        & 63.16\%
        & 346691.87
        & 0.00 \\

        Dense Base Station (8)
        & 90.08\%
        & 68.71\%
        & 346691.87
        & 0.00 \\

        ST-RCP
        & \textbf{93.63\%}
        & \textbf{81.66\%}
        & \textbf{359134.82}
        & \textbf{12442.95} \\

        Dedicated Relay UAV (1)
        & 90.57\%
        & 54.28\%
        & 359407.28
        & 12715.41 \\

        Dedicated Relay UAV (2)
        & 91.81\%
        & 59.64\%
        & 363727.20
        & 17035.33 \\

        Dedicated Relay UAV (3)
        & 93.22\%
        & 66.46\%
        & 378030.53
        & 31338.66 \\

        \hline
    \end{tabular}
\end{table*}

\subsection{Overall Performance Comparison}
\label{subsec:overall_comparison}

This subsection compares ST-RCP with three representative communication strategies: Dense Base Station, FANET, and Dedicated Relay UAV. The comparison is conducted under two one-hour delivery periods extracted from the same-day order data: an off-peak scenario containing 20 orders and a peak scenario containing 90 orders. For ST-RCP, the best-performing selector (GA-CE) is reported for comparison. In the 20-order off-peak scenario, the strongest dedicated-relay configuration with 3 relay UAVs is included as an upper-bound reference, whereas 1 to 3 dedicated relay UAVs are evaluated in the 90-order peak scenario to examine scalability.

\subsubsection{Off-Peak Scenario}

As shown in Table~\ref{tab:offpeak_comparison}, the Dedicated Relay UAV strategy achieves the highest total and descent-phase communication coverage in the off-peak scenario, but incurs an additional energy consumption of $6.95$\~kJ. In comparison, ST-RCP achieves $92.27\%$ total coverage and $66.86\%$ descent-phase coverage with only $1.12$\~kJ of additional energy consumption, demonstrating a more favorable trade-off between communication coverage and energy consumption. Under sparse delivery demand, dedicated relay UAVs maintain an advantage in absolute communication coverage due to their continuous availability, while ST-RCP leverages existing delivery UAVs to substantially reduce the additional energy required for communication recovery.

\subsubsection{Peak Scenario}

Table~\ref{tab:peak_comparison} illustrates the communication performance and energy consumption of different methods in the peak scenario. As shown in Table~\ref{tab:peak_comparison}, ST-RCP with the GA-CE selector achieves a total coverage of $93.63\%$ (comparable to FANET's $93.72\%$ without multi-hop routing overhead) and the highest descent-phase coverage of $81.66\%$. It requires approximately $12.44$~kJ of additional energy, which is lower than the $12.72$~kJ required by a single Dedicated Relay UAV. Increasing the number of Dedicated Relay UAVs improves communication coverage, but at a rapidly increasing energy cost. With three relay UAVs, the descent-phase coverage reaches $66.46\%$, while the additional energy consumption increases to $31.34$~kJ. In comparison, ST-RCP achieves higher communication coverage with substantially lower additional energy consumption. These results indicate that ST-RCP benefits from the larger pool of task-performing UAVs available during peak demand. By opportunistically reusing delivery UAVs as relay resources, ST-RCP achieves a more favorable coverage--energy trade-off than the tested dedicated-relay configurations in the high-demand scenario.

\subsection{Scenario Sensitivity and Applicability}
\label{subsec:sensitivity}

This subsection investigates the effects of infrastructure density and relay-following distance on communication performance, as well as the computational feasibility of ST-RCP.

\subsubsection{Impact of selection strategy}
Table~\ref{tab:objective_ablation_results} summarizes the objective ablation results. For GA, the coverage-per-energy objective achieves the best overall performance, yielding the highest system coverage, the fewest disconnections, the shortest average outage duration, and the lowest additional energy consumption. In contrast, the coverage-per-distance objective achieves the highest descent coverage, while the network-coverage objective provides comparable system coverage but with higher energy consumption. These results highlight the trade-offs between overall connectivity, descent-phase reliability, and resource efficiency. For Greedy, the coverage-per-distance objective yields the highest system coverage, whereas the target-coverage objective achieves the highest descent coverage, the fewest disconnections, and the lowest additional energy consumption.

\begin{table*}[t]
    \centering
    \caption{Communication-oriented objective sensitivity analysis.}
    \label{tab:objective_ablation_results}
    \small
    \begin{tabular}{lcccccc}
        \hline
        \textbf{Optimizer} &
        \textbf{Objective} &
        \textbf{Coverage (\%)} &
        \shortstack{\textbf{Descent}\\\textbf{Coverage (\%)}} &
        \shortstack{\textbf{Disconnect}\\\textbf{Count}} &
        \shortstack{\textbf{Avg. Outage (s)}} &
        \shortstack{\textbf{Additional}\\\textbf{Energy (J)}} \\
        \hline

        \multirow[c]{4}{*}{Greedy}
        & Target Coverage
        & 90.77
        & \textbf{79.16}
        & \textbf{152}
        & 38.26
        & \textbf{16253.95} \\

        & Network Coverage
        & 91.30
        & 77.30
        & 161
        & 36.24
        & 18843.32 \\

        & Coverage per Energy
        & 90.83
        & 75.80
        & 204
        & 38.54
        & 19576.21 \\

        & Coverage per Distance
        & \textbf{91.48}
        & 77.07
        & 197
        & \textbf{35.59}
        & 19713.34 \\

        \hline

        \multirow[c]{4}{*}{GA}
        & Target Coverage
        & 91.56 $\pm$ 0.36
        & 79.24 $\pm$ 1.38
        & 138.0 $\pm$ 23.8
        & 35.45 $\pm$ 1.42
        & 16367.47 $\pm$ 1394 \\

        & Network Coverage
        & 93.36 $\pm$ 1.07
        & 76.90 $\pm$ 1.26
        & 125.3 $\pm$ 7.1
        & 27.15 $\pm$ 4.63
        & 14501.68 $\pm$ 888 \\

        & Coverage per Energy
        & \textbf{93.63 $\pm$ 0.58}
        & 81.66 $\pm$ 0.18
        & \textbf{117.3 $\pm$ 5.8}
        & \textbf{25.61 $\pm$ 2.81}
        & \textbf{12442.95 $\pm$ 1715} \\

        & Coverage per Distance
        & 92.92 $\pm$ 0.08
        & \textbf{82.65 $\pm$ 0.30}
        & 150.3 $\pm$ 27.1
        & 29.09 $\pm$ 0.60
        & 15086.57 $\pm$ 710 \\

        \hline
    \end{tabular}
\end{table*}

\begin{figure*}[t]
    \centering

    \begin{subfigure}[t]{0.31\linewidth}
        \centering
        \includegraphics[width=\linewidth]
        {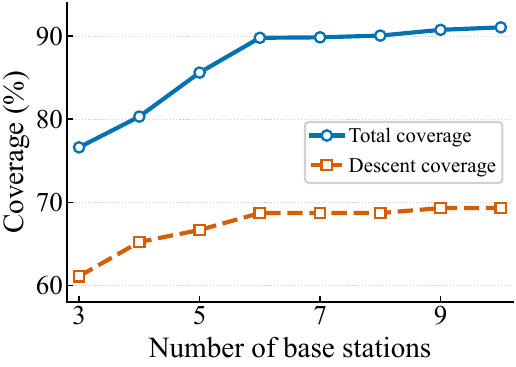}
        \caption{Base-station density.}
        \label{fig:bs_sensitivity}
    \end{subfigure}
    \hfill
    \begin{subfigure}[t]{0.31\linewidth}
        \centering
        \includegraphics[width=\linewidth]
        {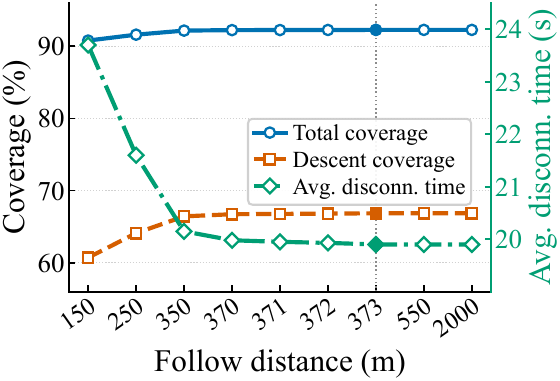}
        \caption{Relay-following distance.}
        \label{fig:follow_distance}
    \end{subfigure}
    \hfill
    \begin{subfigure}[t]{0.31\linewidth}
        \centering
        \includegraphics[width=\linewidth]
        {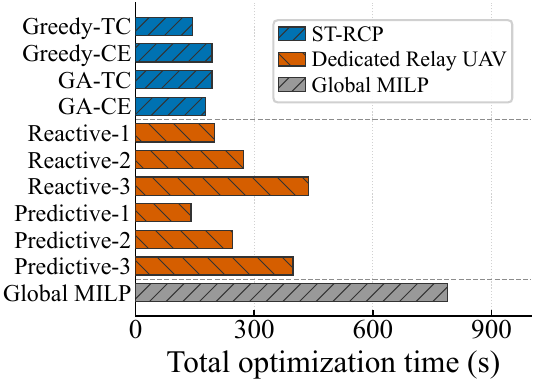}
        \caption{Optimization time.}
        \label{fig:optimization_time}
    \end{subfigure}

    \caption{Sensitivity and computational analysis of ST-RCP:
    (a) impact of base-station density;
    (b) impact of relay-following distance; and
    (c) optimization time of ST-RCP and dedicated-relay strategies.}
    \label{fig:sensitivity_analysis}
\end{figure*}

\begin{figure*}[t]
    \centering
    \begin{subfigure}[t]{0.28\textwidth}
        \centering
        \includegraphics[width=\linewidth]{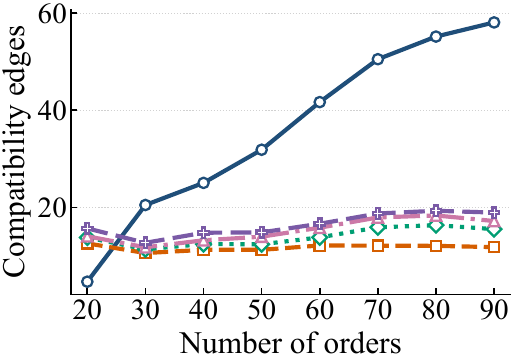}
        \caption{}
        \label{fig:flex_edges}
    \end{subfigure}
    \hspace*{0.6cm}
    \begin{subfigure}[t]{0.28\textwidth}
        \centering
        \includegraphics[width=\linewidth]{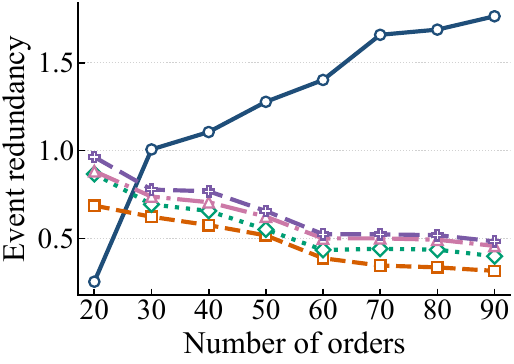}
        \caption{}
        \label{fig:flex_redundancy}
    \end{subfigure}
    \hspace*{0.6cm}
    \begin{subfigure}[t]{0.28\textwidth}
        \centering
        \includegraphics[width=\linewidth]{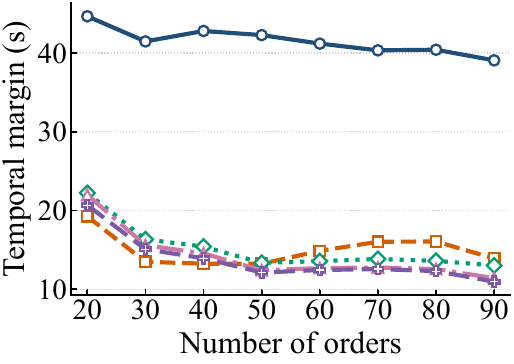}
        \caption{}
        \label{fig:flex_margin}
    \end{subfigure}

    \vspace{0.4em}
    \includegraphics[width=0.55\textwidth]{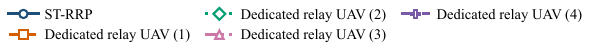}

    \caption{Process-flexibility comparison:
    (a) Compatibility-edge count;
    (b) Event redundancy; and
    (c) Temporal margin.}
    \label{fig:process_flexibility}
\end{figure*}

\subsubsection{Impact of base-station density}
As shown in Fig.~\ref{fig:bs_sensitivity}, increasing the number of base stations from three to six substantially improves both total and descent-phase coverage, while further densification yields only marginal gains. This indicates diminishing returns from infrastructure expansion and shows that additional base stations cannot fully eliminate building-induced communication blind spots.

\subsubsection{Impact of relay-following distance}
Fig.~\ref{fig:follow_distance} shows that increasing the maximum relay-following distance from $150$~m to approximately $373$~m improves coverage and reduces average disconnection time, after which the performance becomes nearly stable. Thus, most relay benefits can be achieved within a moderate following-distance range.

\subsubsection{Computational applicability}
As shown in Fig.~\ref{fig:optimization_time}, the ST-RCP selectors require $143$--$234$~s of optimization time, whereas the dedicated-relay strategies increase to $398$--$438$~s with three relay UAVs. In contrast, the global MILP incurs the highest computational cost, highlighting the computational burden of global joint optimization. Although the single-relay predictive baseline is slightly faster than the fastest ST-RCP configuration, ST-RCP maintains a relatively stable computational cost across its evaluated selectors, providing a favorable balance between communication performance and computational efficiency.

\subsection{Mechanism and Process-Flexibility Analysis}
\label{subsec:mechanism}

This subsection evaluates how ST-RCP converts task-performing delivery UAVs into feasible relay alternatives through compatibility-graph configuration and departure scheduling. In the evaluation, pre-departure shifts in Eq.~\eqref{eq:flexibility_scheduling} are searched over a discrete grid within $[0,\bar{\delta}_h]$ ($10\,\mathrm{s}$ step) under nominal operation and single-helper unavailability scenarios $\Omega$. The resulting structural adaptability is quantified using the three process-flexibility metrics defined in Eq.~\eqref{eq:flexibility_metrics}: mean compatibility-edge count $\bar{A}$, mean event redundancy $\bar{R}$, and mean temporal margin $\bar{M}$. Fig.~\ref{fig:process_flexibility} shows how these metrics evolve as the number of delivery orders increases.

\begin{figure*}[t]
  \centering
  \includegraphics[width=0.975\textwidth]{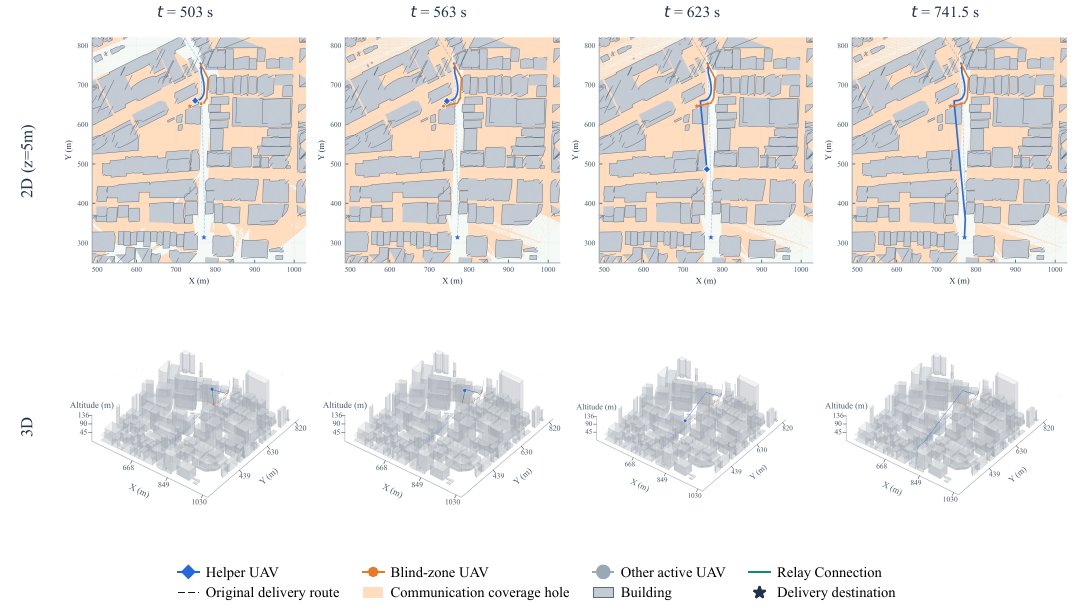}
  \caption{Spatiotemporal evolution of a proactive relay episode under the 90-order peak scenario at $t = 503$, $563$, $623$, and $741.5$\,s. 
  The helper UAV (blue) proactively establishes and maintains the relay link (green) for the blind-zone UAV (orange) within the coverage hole, and subsequently resumes its nominal delivery route upon task completion.}
  \label{fig:relay_timeseries}
\end{figure*}

\subsubsection{Compatibility Structure and Event Redundancy}

Fig.~\ref{fig:flex_edges} compares the mean compatibility-edge count $\bar{A}$, i.e., the average number of feasible helper--event relationships available at each evaluation epoch. Under ST-RCP, $\bar{A}$ increases substantially with delivery demand, indicating that a denser delivery fleet creates a richer set of feasible relay relationships. In contrast, the dedicated-relay schemes exhibit only a moderate increase because their relay resource pools remain fixed.
Fig.~\ref{fig:flex_redundancy} further reports the mean event redundancy $\bar{R}$, which measures the average number of feasible alternative helpers available for each active blind-zone event. Under ST-RCP, $\bar{R}$ increases with delivery demand, whereas it generally decreases for the dedicated-relay schemes. Thus, although higher demand introduces more communication events, it also supplies additional task-performing UAVs that can serve as alternative helpers. This explains why ST-RCP becomes increasingly effective in denser delivery operations.

\subsubsection{Temporal Flexibility}

Fig.~\ref{fig:flex_margin} compares the mean temporal margin $\bar{M}$ of feasible helper--event relationships. ST-RCP consistently achieves a substantially larger temporal margin than the dedicated-relay strategies across different order volumes. Although $\bar{M}$ gradually decreases as demand increases, it remains around $40$~s in the densest scenario, whereas the dedicated-relay configurations retain only about $11$--$14$~s.

Overall, the three metrics reveal the mechanism underlying ST-RCP. Increasing delivery demand enlarges the feasible compatibility set, increases the number of alternative helpers available to each event, and still preserves comparatively large temporal slack. In contrast, a fixed dedicated-relay fleet must accommodate increasing relay demand with a limited resource pool. The scalability benefit of ST-RCP therefore arises from its ability to convert additional task-performing UAVs into feasible and temporally viable relay alternatives.

\subsection{Spatiotemporal Evolution of a Relay Cooperation}
\label{subsec:episode}

Finally, Fig.~\ref{fig:relay_timeseries} illustrates a representative relay cooperation episode under the peak demand scenario. At $t=503~\mathrm{s}$, the helper UAV departs from its nominal delivery route and proceeds to the designated relay position, where it establishes and maintains a relay link with the UAV operating in the communication blind spot to ensure communication continuity. Upon completion of the assisted UAV's mission in the blind spot, the helper UAV resumes its original delivery route to fulfill its primary task.

\section{Conclusion}

This paper proposed a predictive spatio-temporal relay cooperation and planning (ST-RCP) framework that leverages task-performing delivery UAVs as opportunistic communication relays in complex urban environments. By integrating communication blind-spot prediction, helper--event coordination, constraint-aware trajectory replanning, and process flexibility analysis, ST-RCP enhances communication continuity while reducing reliance on dedicated relay resources. The results further demonstrate that the effectiveness of relay cooperation depends on the availability, redundancy, and temporal flexibility of feasible helper resources.
Despite these promising results, the current study is limited to simulation-based evaluations under selected urban scenarios, with precomputed communication maps and deterministic trajectory predictions. Future work will extend the framework to more diverse urban environments, incorporate uncertainties in communication conditions and UAV trajectories into online planning, and further validate the proposed framework through larger-scale simulations, repeated experiments, and real-world or hardware-in-the-loop testing.

\bibliographystyle{IEEEtran}
\bibliography{references}
\vskip 0pt plus -1fil
\begin{IEEEbiography}[{\includegraphics[width=1in,height=1.25in,clip,keepaspectratio]{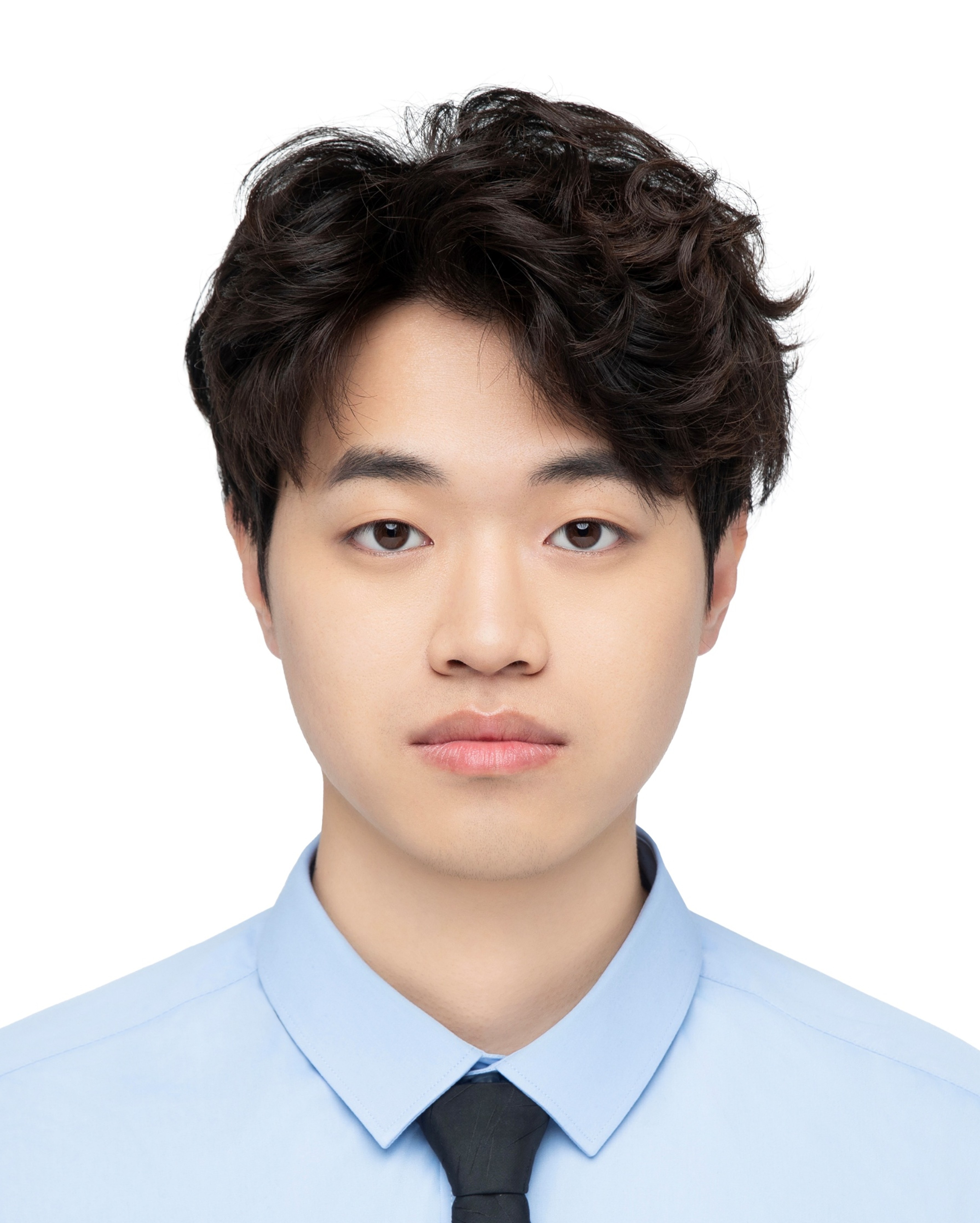}}]{Haoyu Zhang}
received the B.Sc. and M.Sc. degrees from Zhejiang University, Hangzhou, China, in 2023 and 2026, respectively. He is currently pursuing the Ph.D. degree with the Department of Mechanical and Aerospace Engineering, The Hong Kong University of Science and Technology, Hong Kong, China. His research interests include UAV operations, multi-UAV coordination, and reinforcement learning.
\end{IEEEbiography}
\vskip 0pt plus -1fil
\begin{IEEEbiography}[{\includegraphics[width=1in,height=1.25in,clip,keepaspectratio]{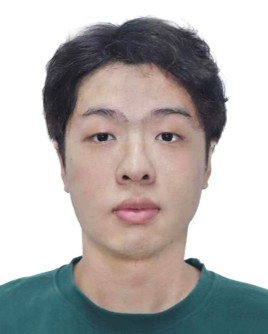}}]{Yingjie He}
received the B.Sc. degree from the Harbin Institute of Technology (Shenzhen), Shenzhen, China. He is currently pursuing the Ph.D. degree with the Department of Mechanical and Aerospace Engineering, The Hong Kong University of Science and Technology, Hong Kong, China. His research interests include UAV operations, multi-UAV coordination, low-altitude infrastructure, and reinforcement learning.
\end{IEEEbiography}
\vskip 0pt plus -1fil
\begin{IEEEbiography}[{\includegraphics[width=1in,height=1.25in,clip,keepaspectratio]{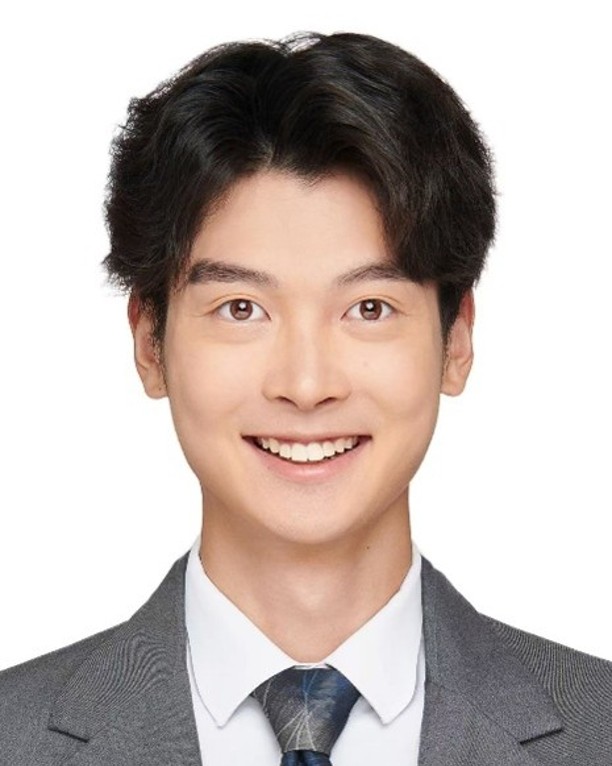}}]{Zehao Zhuo}
received the B.Sc. degree from Zhejiang University, Hangzhou, China, in 2025. He is currently pursuing the Ph.D. degree with the College of Civil Engineering and Architecture, Zhejiang University, Hangzhou, China. His research interests include low-altitude transportation systems, air--ground cooperative transportation, transportation system optimization, and robust optimization.
\end{IEEEbiography}
\vskip 0pt plus -1fil
\begin{IEEEbiography}[{\includegraphics[width=1in,height=1.25in,clip,keepaspectratio]{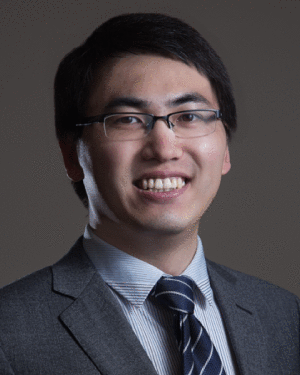}}]{Xiqun (Michael) Chen}
received the B.E. and Ph.D. degrees from the Department of Civil Engineering, Tsinghua University, Beijing, China, in 2008 and 2013, respectively. He is currently a tenured Professor at Zhejiang University, Hangzhou, China, and the Director of the Institute of Intelligent Transportation Systems. He has published more than 200 peer-reviewed journal articles and over 100 conference papers. His research interests include intelligent transportation systems, artificial intelligence for transportation, transportation simulation and optimization, transportation big data analytics, and shared mobility. He has received the National Distinguished Young Scholars Award from the National Natural Science Foundation of China, the Science and Technology Innovation Youth Award from the China Communications and Transportation Association, the Science and Technology Award from the China Intelligent Transportation Systems Association, the 2013 IEEE Intelligent Transportation Systems Society Best Ph.D. Dissertation Award, and more than ten best paper awards at international conferences. He serves as an Associate Editor of \emph{Transportation Research Part C: Emerging Technologies} and \emph{IEEE Transactions on Intelligent Vehicles}, an Academic Editor of \emph{The Innovation}, and an Editorial Advisory Board Member of \emph{npj Sustainable Mobility and Transport}.
\end{IEEEbiography}
\vskip 0pt plus -1fil
\begin{IEEEbiography}[{\includegraphics[width=1in,height=1.25in,clip,keepaspectratio]{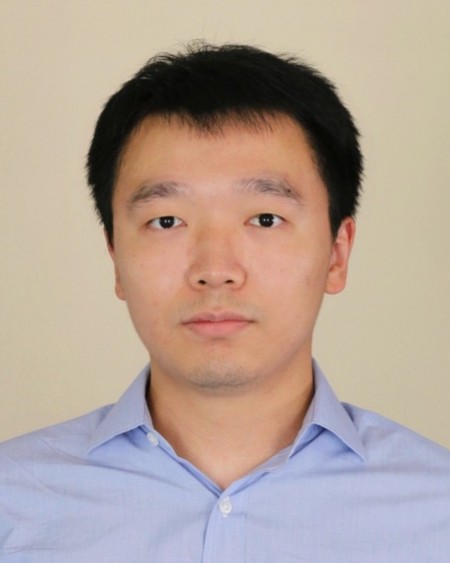}}]{Zhenyu Gao}
is an Assistant Professor of Mechanical and Aerospace Engineering at The Hong Kong University of Science and Technology (HKUST). He holds a Ph.D. in Aerospace Engineering and an M.S. in Operations Research from the Georgia Institute of Technology, as well as a B.S. in Aerospace Engineering with a minor in Mathematics from the University of Illinois Urbana-Champaign. His Ph.D. dissertation was recognized with the 2023 Georgia Tech Sigma Xi Best Ph.D. Thesis Award. Before joining HKUST, he was a postdoctoral fellow at The University of Texas at Austin. His research focuses on aerospace systems design and optimization, autonomous systems, and data intelligence. 
\end{IEEEbiography}

\end{document}